\documentclass{article}

\PassOptionsToPackage{numbers,compress}{natbib}

\usepackage[preprint,eandd]{neurips_2026}
\usepackage{graphicx}
\usepackage{amsmath}
\usepackage{wrapfig}

\usepackage[utf8]{inputenc} % allow utf-8 input
\usepackage[T1]{fontenc}    % use 8-bit T1 fonts
\usepackage{hyperref}       % hyperlinks
\usepackage{url}            % simple URL typesetting
\usepackage{booktabs}       % professional-quality tables
\usepackage{amsfonts}       % blackboard math symbols
\usepackage{nicefrac}       % compact symbols for 1/2, etc.
\usepackage{microtype}      % microtypography
\usepackage[table]{xcolor}  % colors
\usepackage{pifont}         % \ding for check/cross marks
\usepackage{placeins}       % \FloatBarrier to keep tables/figures within sections
\usepackage{array}          % >{\centering\arraybackslash} for centered p-columns
\definecolor{CSBGreen}{HTML}{1B7F5F}
\definecolor{CSBRed}{HTML}{B33A4B}
\definecolor{CSBPanel}{HTML}{E9EFF6}
\definecolor{CSBPanelRule}{HTML}{D7E1EC}
\definecolor{CSBInk}{HTML}{253246}
\definecolor{CSBMuted}{HTML}{5F6B7A}
\newcommand{\cmark}{\textcolor{CSBGreen}{\ding{51}}}
\newcommand{\xmark}{\textcolor{CSBRed}{\ding{55}}}
\newcommand{\pmark}{\textcolor{CSBMuted}{$\triangle$}}
\newcommand{\casebox}[1]{%
  \par\vspace{0.25em}
  \noindent\begingroup
  \setlength{\fboxsep}{5pt}%
  \colorbox{CSBPanel}{\parbox{\dimexpr\linewidth-2\fboxsep\relax}{\small #1}}%
  \endgroup
  \par\vspace{0.25em}
}
\graphicspath{{figures/}}

\newcommand{\benchmark}{\textsc{ConsumerSimBench}}
\newcommand{\model}{\textsc{ConsumerSCF}}
\newcommand{\avg}{\textsc{Avg}}
\makeatletter
\renewcommand{\@noticestring}{Preprint. Dataset: \url{https://huggingface.co/datasets/wty500/ConsumerSimBench}.}
\makeatother

\title{Can LLMs Think Like Consumers?\\
Benchmarking Crowd-Level Reaction Reconstruction with \benchmark{}}

\author{%
Tianyu Wang\\
Shanghai Jiao Tong University\\
\texttt{wty500@sjtu.edu.cn}
\And
Jiajun Li\\
Noumena AI\\
\texttt{taringlee@gmail.com}
\And
Jianghao Lin\\
Shanghai Jiao Tong University\\
\texttt{linjianghao@sjtu.edu.cn}
}

\begin{document}

\maketitle
\begin{abstract}
LLMs are increasingly used as ``digital consumers'' to simulate public opinion, pre-test marketing decisions, and anticipate audience response. However, existing evaluations rarely ask whether a model can reconstruct the concrete reaction patterns that real consumers surface in public discourse. We introduce \benchmark{}, a benchmark built from 1{,}553 real Chinese social-media topics and 23{,}122 atomic, rule-audited criteria spanning four reaction families. Rather than scoring open-ended generations with a holistic preference judge, \benchmark{} decomposes each task into auditable yes-no decisions over concrete reaction points, raising three-judge agreement from 65.8\% to 92.1\% with 98.4\% agreement between pointwise judge decisions and human-majority labels. Across 13 frontier generators, the strongest model, Gemini-3.1-Pro, covers only 47.8\% of real reaction criteria, while GPT-5.2 and Claude-4.6 trail far behind despite their strength on technical benchmarks. The failures reveal a sharp gap between technical-benchmark performance and socially grounded consumer intuition. A direct structured reasoning prompt decreases coverage, while a generate--reflect multi-agent pipeline improves MiMo-V2.5-Pro from 32.9\% to 37.6\% on a subset. \benchmark{} reframes consumer simulation as a forecasting problem over real public-discourse reactions, showing that frontier LLMs remain far from reliably predicting what consumers will actually care about in high-context Chinese consumer discourse.
\end{abstract}

\begin{figure*}[h]
\vspace{-1em}
    \centering
    \includegraphics[width=\textwidth]{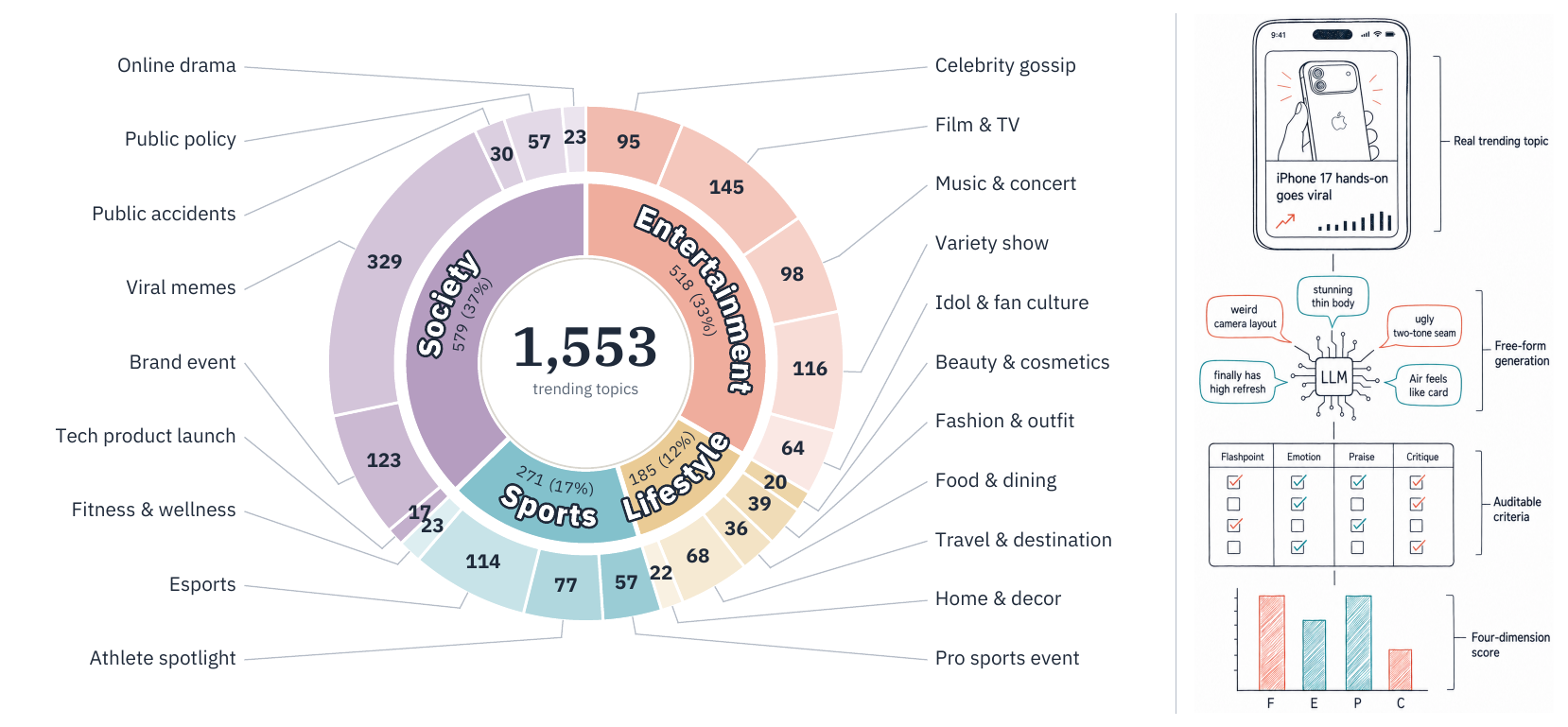}
    \vspace{-2em}
    \caption{\textbf{Overview of \benchmark{} .} \emph{Left:} The 1{,}553 trending topics span four super-categories and twenty consumer-facing sub-fields. \emph{Right:} One task instance. Given a real trending topic, a generator must produce a free-form bundle of consumer comments that collectively cover an audited set of atomic criteria across four reaction families (flashpoints, emotion, praise, critique).}
    \vspace{-1em}
    \label{fig:hero}
\end{figure*}

\section{Introduction}\label{sec:intro}
\begin{center}
  \vspace{-1ex}
  \fbox{%
    \parbox{0.75\linewidth}{\centering\small\itshape
      ``If an LLM can truly think like a consumer,\\
      it should be able to predict what a real user would say about a brand.''
    }
  }
  \vspace{-1ex}
\end{center}
The vision of LLMs acting as \emph{digital consumers} is rapidly moving from demos to deployment. Teams increasingly use LLM agents to simulate social opinion \citep{yang2024social, piao2025agentsociety, zhang2025trendsim}, pre-test marketing messages \citep{wang2025user, chu2025llm}, and design social strategy \citep{mou2024individual,qu2024performance}. The motivation is straightforward: traditional consumer research is slow and expensive, while large-scale scalar signals such as click-through rate (CTR) are post-hoc, platform-specific, and often fail to explain \emph{why} consumers react the way they do.
% Large Language Model (LLM)-based agents are increasingly deployed to perform tasks that were traditionally carried out by humans, including public opinion mocking \citep{yang2024social, piao2025agentsociety, zhang2025trendsim}, user research \citep{wang2025user, chu2025llm}, social strategy design \citep{mou2024individual}, and applications in psychology and behavioral studies \citep{xie2024can, hewitt2024predicting, anthis2025llm, zakazov2024assessing, xiao2025evaluating}.
% Meanwhile, a growing body of work evaluates LLMs' social reasoning and Theory of Mind (ToM), showing strong performance on classic false-belief style tests under controlled settings \citep{kosinski2023tom}.

However, downstream decision-makers still lack a principled understanding of \textbf{how closely LLM simulations align with actual human behaviors} in high-context, strategic, and socially intensive environments \citep{wang2025llm}.
For example, on lifestyle-oriented user-generated-content (UGC) platforms such as RedNote or Youtube, creators often rely on pragmatic moves such as reverse signaling, humblebragging, and anxiety induction to shape how audiences interpret a post.
In such settings, surface-level understanding (e.g., \emph{the text expresses sadness}) is insufficient; what matters is \textbf{strategic social intelligence} (e.g., \emph{the YouTuber creates a sad atmosphere for his advertisement}), which requires understanding the narrative stance and how it is meant to shape the audience's inference. This can fundamentally reshape the marketing industry, because reliably predicting consumer reactions before publication could defuse the next viral crisis instead of apologizing for it, and identify breakout angles before chasing them, which remains an extremely difficult mission even for marketing experts.

Existing benchmarks are inadequate for three reasons.
\textbf{First}, real UGC does not follow the logic in standard ToM-style tests. ToM examples are typically built around a strict reasoning trace (e.g.\ \emph{I stole the hat. He did not see. So, he doesn't know the hat is missing.}) \citep{chen2025theory}, \textbf{where the key facts are explicit and the answer space is closed}. In contrast, real UGC's meanings can be layered, implicit, and involve satire or boasting (e.g.\ \emph{It is so annoying to decide which house to decorate!}).
\textbf{Second,} many benchmarks emphasize the prediction of a choice, distribution, scalar number, or fixed user trace (e.g., CTR, SSR \citep{10.1145/3701716.3715258, maier2025llms}, public-opinion distributions \citep{miranda2025simulating}, social-media engagement labels and conditioned comment prediction \citep{bojic2026llm, schwager2026towards}, popularity prediction \citep{SMP2023}, etc.), which cannot evaluate models' open-ended humanities-side capacity \citep{hu2025simbench}. Practitioners care about \emph{what real people will say and why}; most benchmarks only measure \emph{what label a model predicts}.
% leaving it unclear whether a model is genuinely simulating human behavior or merely ``guessing'' the ground-truth distribution from training-data 
% 评测；方法；Application
\textbf{Third,} many evaluations rely on an LLM to directly assign a holistic final score for an open-ended output, which can be subjective, unstable, and vulnerable to systematic biases \citep{tu2024charactereval, liu2024roleagent, gu2024llmasajudge, chen-etal-2024-humans}.
This makes it difficult to interpret what a numeric score means and whether it is reliable across models, prompts, and domains.

Meanwhile, despite LLMs' top performance on quantitative leaderboards \citep{wang2024mmlu, jimenez2023swe, merrill2026terminal}, \textbf{social-media users continue to complain that these models are weak at the ``humanities'' side of language}---reading people, registering social nuance, and showing human touch. Part of this perceived gap is a benchmarking artifact, where very few benchmarks audit a model's humanities-side skills.
% Addressing these limitations involves several challenges.
% \textbf{First,} ``human-likeness'' in high-context UGC is inherently ambiguous and difficult to formalize.
% \textbf{Second,} even with a workable definition, it is non-trivial to design \emph{automatic} evaluation metrics that are reliable, simple, stable, and aligned with the definition without collapsing into another LLM-as-a-Judge setup.
% \textbf{Third,} suitable data sources must be high-quality while avoiding copyright and licensing risks.
% \textbf{Finally,} a practically useful benchmark should generalize across categories and user groups, and it should provide actionable insights for real-world deployment rather than a single opaque score.

To bridge these gaps, we introduce \benchmark{}, a benchmark designed to evaluate whether LLMs can simulate \textbf{authentic consumer behaviors} in the wild.
We use RedNote as a major consumer-facing UGC platform where real reactions to products, brands, entertainment releases, lifestyle trends, and public events are observable at scale. Its dense, high-context consumer discourse makes it a representative and practically important setting for testing whether models can anticipate what consumers will notice, praise, attack, and emotionally amplify.
Concretely, given a trending topic and its event description, a model is required to generate realistic comments. We audit coverage over four reaction families: \textbf{(i)} \emph{sentiment flashpoints}, the concrete social triggers; \textbf{(ii)} \emph{emotion keywords}, the overall sentiment of the real crowd; \textbf{(iii)} \emph{positive aspects}, what users praise; \textbf{(iv)} \emph{negative aspects}, what users attack. Every criterion is atomic, defined, example-anchored, and rule-audited. We find that even the strongest model (Gemini-3.1-Pro: 47.8\%) performs poorly on this humanities task, and top-tier models that look comparable on technical benchmarks diverge sharply on \benchmark{} (GPT-5.2: 35.8\%, Claude-Opus-4.6: 32.8\%). This gives measurable feedback on a humanities-side capability that current LLM training and evaluation often underemphasize.
% Concretely, \benchmark{} evaluates an agent's ability to generate a user's likely response to a brand-related stimulus conditioned on that user's historical UGC, and to do so in a way that matches both the \emph{observed outcome} and the \emph{latent pragmatic intent} of real human posts. Our key technical idea is that success in such environments requires \emph{Strategic ToM}, and therefore propose \model{}, a structured Social Cognition Framework, to effectively decompose the latent social meaning of UGC and achieve SOTA performance on \benchmark{}.
\newpage
\textbf{Our main contributions are:}
\begin{itemize}
    \item To study whether an LLM can \emph{truly think like a consumer}, we propose \textbf{\benchmark{}}, a UGC-behavior simulation benchmark that operationalizes this question as \textbf{crowd-level reaction reconstruction}: given a real trending topic and event description, can the model reconstruct the reaction patterns that real consumers surfaced in public discourse?
    \item To make open-ended consumer-likeness measurable, \benchmark{} contains 1,553 real topic instances and 23,122 atomic scoring criteria that jointly evaluate emotion keywords and emotional bursts \textbf{at the macro level}, and content-faithful expressions of individual consumer feedback \textbf{at the micro level}.
    \item For automatic evaluation, we construct detailed criteria that turn open-ended scoring from holistic ``judge the answer'' decisions into \textbf{auditable yes-no questions}. This yields 92.1\% three-judge agreement, reaches 98.4\% agreement with human-majority labels, and outperforms holistic LLM-as-Judge scoring (65.8\%) and traditional similarity metrics such as Jaccard and N-gram overlap.
    \item We test both a structured-prompt and a generate--reflect multi-agent pipeline. A direct \model{} prompt hurts, while iterative multi-agent refinement improves MiMo-V2.5-Pro by +4.7 points and GPT-5.2 by +1.8 points on a 100-topic slice.
    % \item To validate the effectiveness of \benchmark{}, we propose \textbf{\model{}}, a ToM-grounded modeling approach that predicts SCF factors and conditions generation on them, which achieves SOTA performance on \benchmark{}; \textbf{we further apply \model{} to real markering scenarios}.
\end{itemize}

\section{Related Work}
% We position \benchmark{} and the related work by \emph{unit of inference}, shown  in Table~\ref{tab:positioning}.

\begin{table*}[t]
\centering
\caption{Positioning by \emph{unit of inference}. Consumer simulation requires open-form, crowd-level reconstruction of real public discourse with auditable targets. Among the benchmark families summarized here, \benchmark{} is the only one that satisfies all five criteria jointly.}
\label{tab:positioning}
\small
\resizebox{\textwidth}{!}{%
\begin{tabular}{@{}lccccc@{}}
\toprule
\rowcolor{CSBPanel}
\textbf{Benchmark family} & \textbf{Unit of inference} & \textbf{Real public discourse} & \textbf{Free-form generation} & \textbf{Strategic targets} & \textbf{Auditable eval} \\
\midrule
ToM \citep{kim2023fantom, xu2024opentom, chen2025theory} & Dyad (Alice\,/\,Bob) & \xmark\ (fictional) & \xmark\ (MCQ) & \pmark & \pmark \\
CTR (scalar prediction) & Aggregate scalar & \cmark & \xmark & \xmark & \cmark \\
Persona Simulation \citep{Park2023GenerativeAgents, wang2025user} & Individual trajectory & \pmark & \cmark & \pmark & \xmark \\
\textbf{\benchmark{}(Ours)} & \textbf{Crowd reaction} & \cmark & \cmark & \cmark & \cmark \\
\bottomrule
\end{tabular}%
}
\vspace{-2em}
\end{table*}
% \paragraph{Dyadic belief inference (ToM).} FanToM, OpenToM, Hi-ToM and related benchmarks \citep{kim2023fantom, xu2024opentom, chen2025theory} track what one agent believes another believes in stylized, closed-form, typically multiple-choice scenarios. The unit is a dyad; the setting is fictional. They illuminate a cognitive competence but do not ask whether an LLM can reconstruct what a \emph{crowd} will say to a \emph{real public event}.

\paragraph{Persona simulation.} Recent work has explored using LLMs as proxies for human behavior in opinion, market, and psychological studies \citep{pmlr-v202-aher23a, lin2026large, qi2025cross, ludwig2026extracting, wang2026large}. Generative Agents \citep{Park2023GenerativeAgents} demonstrated that LLM-powered agents with memory, planning, and reflection can exhibit emergent social behaviors in a simulated town. Subsequent work has extended this paradigm to social-science experiments \citep{hewitt2024predicting, anthis2025llm, zakazov2024assessing, xiao2025evaluating, slumbers2025using}, user-behavior simulation \citep{wang2025user, chu2025llm, yang2024social, piao2025agentsociety, zhang2025trendsim, yang2024oasis, zhou2023sotopia}, cognitive reasoning and human-response modeling \citep{zhou2025think, hwang2025infusing, binz2025foundation}, customer simulation \citep{wang2025customer, wangcustomer}, marketing simulation \citep{brand2023using, agarwal2025silicon, yang2025twinmarket, li2024econagent, yu2024fincon, su2026sell}, opinion dynamics and pluralism \citep{chuang2025debate, poole2025benchmarking}, multi-turn user simulation \citep{dou2025simulatorarena, seshadri2026lost}, war simulation \citep{hua2023war}, comprehensive systems and agentic public-opinion analysis \citep{gao2023s3, tang2025gensim, zhang2025socioverse, chen2024agentverse, wang2025megaagent, liu2025can}, and others \citep{gao2024large, chenpersona}. Critical studies also warn that synthetic respondents can fail systematically as replacements for human survey data \citep{bisbee2024synthetic}. Across these works, the similarity between LLM agents and real human social behavior is mostly evaluated through classical psychology, survey, or social-science experiments, which risk data contamination or being solved by general common sense rather than genuine open-ended reaction reconstruction.

\paragraph{Evaluating LLMs' mindset.} Current benchmarks mainly rely on multiple-choice \citep{sap2019social, lee2025llms, zeng2025psychcounsel, chen2024socialbench, marraffini2024greatest, anonymous2025socialr, yong2025motivebench, sabour2024emobench, kang2025hssbench}, classification \citep{hendrycks2021aligning}, LLM-as-a-Judge \citep{shao2023character, tu2024charactereval, wang2024incharacter, samuel-etal-2025-personagym, Klinkert_Buongiorno_Clark_2024, ong2025human, mou-etal-2025-agentsense}, and similarity metrics like Rouge-L \citep{wang-etal-2024-rolellm} and BERTScore \citep{emelin2021moral}. Beyond emotional and cognitive benchmarks \citep{chen2024emotionqueen, liu2024interintent, gandhi2023understanding, echterhoff2024cognitive, jia2024decision}, a parallel line of work uses Theory-of-Mind (ToM) tasks \citep{chen2025theory} to track what one agent believes another believes, in stylized, closed-form, typically multiple-choice scenarios \citep{shinoda2025tomato, kim2023fantom, xu2024opentom, strachan2024testing, wu2023hi, shi2025muma, liu2025mind, zhang2024affective}. While these benchmarks illuminate a cognitive competence, they test reasoning ability rather than the actual emulation of human social behavior, and their scoring is largely subjective. Our work advances benchmarking on humanities-side capabilities with auditable criteria even when the answer is not unique.

% \paragraph{LLM-as-judge.} Holistic or pairwise LLM judges \citep{gu2024llmasajudge, chen-etal-2024-humans} are practical but unstable without decomposition. Our methodological move is to turn open-form simulation into a sequence of pointwise binary questions with definitions, examples, and rules; Section~\ref{sec:judge} shows this pushes three-judge agreement from 65.8\% (batch) to 92.1\% (semantic-rules pointwise).

\paragraph{Position.} As shown in Table~\ref{tab:positioning}, \benchmark{} targets a different competence from ToM. ToM asks whether a model can \emph{\textbf{reason}} about a mind; \benchmark{} asks whether a model can \emph{\textbf{reconstruct}} what a crowd will think, feel, and say about a real brand event.

\section{\benchmark{} Dataset Design}
\subsection{Design goals} \benchmark{} targets four design goals. \textbf{1. Authenticity over generic plausibility:} Rather than synthesizing diverse scenarios, we ground every record in real public reactions. Models must reconstruct the salient reaction points real consumers surfaced about a real brand, product, or public event, not merely sound emotional; this evaluates public-trend reaction coverage rather than median survey sentiment. \textbf{2. Free-form outputs:} A model should not merely classify sentiment into a pre-set category \citep{argyle2023out}; it should simulate what a crowd of consumers would plausibly say. \textbf{3. Derived artifacts:} For responsible release, we publish derived, anonymized artifacts rather than raw user histories or identifiers. \textbf{4. Auditable scoring:} Instead of asking a vague \emph{``How good is this answer?''} question like what LLM-as-Judge benchmarks do (``40\% human-like'' is meaningless), we change it into a series of auditable decisions: \emph{``Did the generated comments cover this particular reaction point?''}, where every point is an inspectable criterion with definitions, examples, and rules. The result is far more stable and easier to interpret than a global similarity score.

\begin{figure*}[t]
    \centering
    \includegraphics[width=0.8\textwidth]{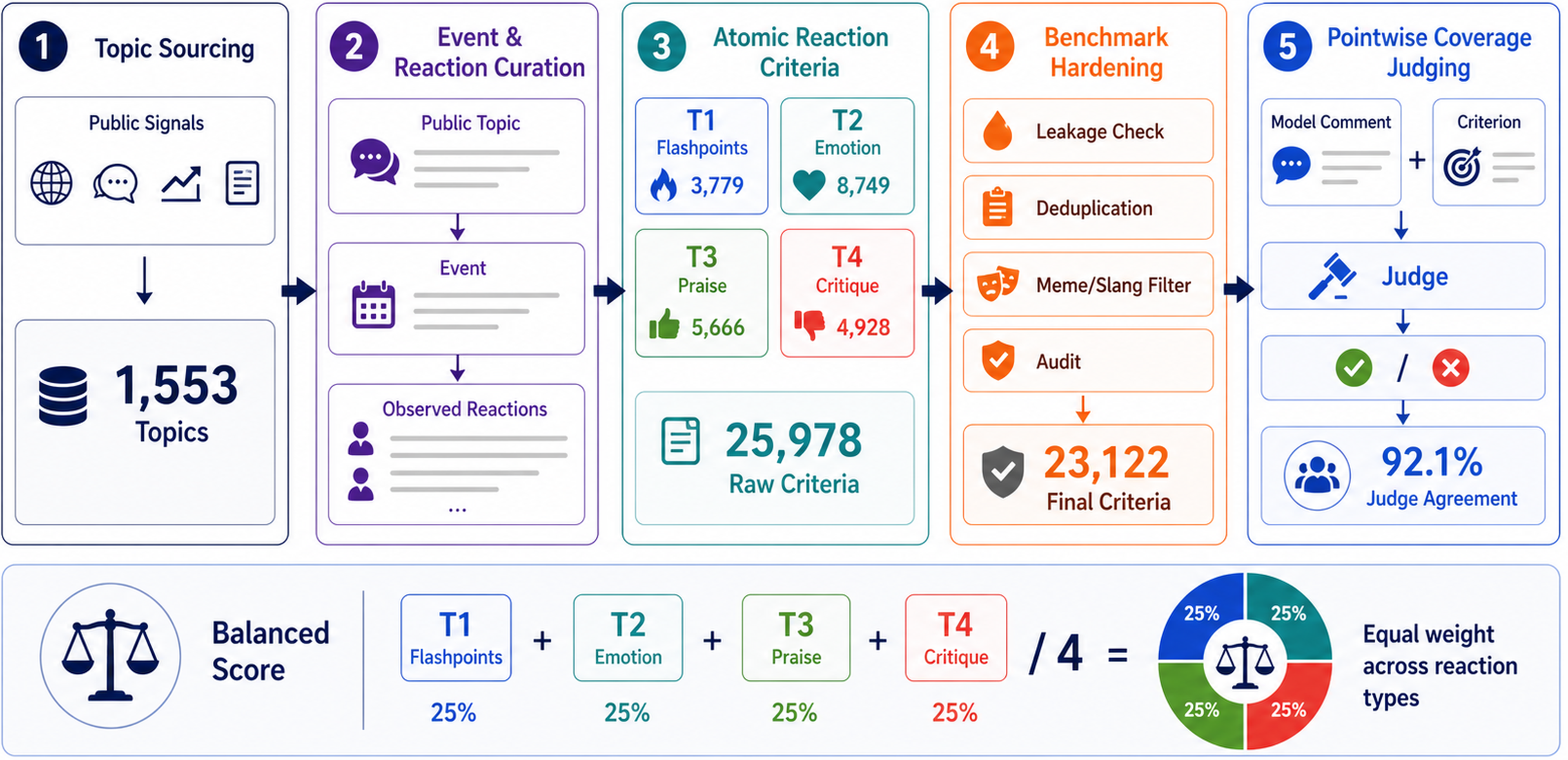}
    \caption{\textbf{\benchmark{} construction and evaluation pipeline.} Public trend signals are curated into topic--event records and abstracted observed reactions; these reactions are converted into four families of atomic reaction criteria, hardened, and finally used for pointwise coverage judging of model-generated comments. The final score gives equal weight to the four reaction families.}
    \label{fig:methodology_overview}
    \vspace{-0.75em}
\end{figure*}

\subsection{Task formulation}\label{sec:task-formulation} Each topic has a topic keyword $k$ and a neutral event description $d$. Given the $i$-th topic $\{k_i,d_i\}$, an LLM $\pi$ is required to produce $n_i$ (10--30) realistic comments. The comments will be graded from four dimensions, each contributing $25\%$ of the total score:
\begin{itemize}
    \item \textbf{T1 sentiment flashpoints}: The emotional trigger that real consumers resonate with, which can be one or many phrases. This is usually the reason that pushes the topic onto the trending list. Accurately predicting these triggers, both positive and negative, can help brands avoid reputation crises while identifying possible viral hits. For example, the topic \emph{Arc'teryx's fireworks} has a flashpoint named \emph{environmental protection}, where people's concern about the fireworks' environmental impact later led to a public-relations crisis. The grading process judges how many flashpoints are clearly identified, and the score is the ratio.

    \item \textbf{T2 emotion keywords}: The affective vocabulary that best describes people's attitude in the comments (such as \emph{resentful}, \emph{conflicted}, \emph{put on the spot}). They can cover many aspects, and all the words are open-form. There are no preset categories. The grading process identifies how many emotions are present in the generated posts, using semantic coverage rather than exact word matching.

    \item \textbf{T3 positive aspects:} The specific \emph{details} appearing in the original event that consumers like, praise, or upvote (such as \emph{cost-effectiveness} and \emph{emotional value}). The marketing industry can use this to test if their proposed selling point will lead to the intended response.

    \item \textbf{T4 negative aspects:} Similar to T3, these are the details on which criticism occurs against the original event (such as \emph{discrimination} or \emph{affectation}). A capable model on this can help the marketing industry avoid inappropriate phrasing in their post.
\end{itemize}

Within each section, the score is
\begin{equation}
    s_{i,j}=\frac{\# \text{ of matched criteria in }T_j}{\# \text{ of total criteria in } T_j}
\end{equation}
and the final total score is
\begin{equation}
    s=\frac{1}{N}\sum_{i=1}^N\frac{s_{i,1}+s_{i,2}+s_{i,3}+s_{i,4}}{4}.
\end{equation}

\paragraph{Atomic Criteria.}
Each of the four dimensions decomposes into atomic criteria that can be judged independently. As shown in Table~\ref{tab:criterion_example}, every atomic criterion contains five fields: \textbf{1.} the corresponding trigger / keyword / aspect; \textbf{2.} a detailed definition; \textbf{3.} positive examples that meet the criterion; \textbf{4.} negative examples that do not; \textbf{5.} a one-sentence judgment rule that disambiguates edge cases.  A single comment may cover multiple criteria, and evaluation is over the concatenated comment set rather than one-to-one post matching. 

For each atomic criterion, we ask a judge model whether the generated comments cover that criterion. To keep evaluation fair across generators that produce different numbers or lengths of comments, we concatenate each model's comments and truncate the judge-visible text to 3{,}600 characters, the estimated empirical 99\% coverage bound, and adapt $n_i$ to follow the empirical comment-volume. The judge receives all the above criteria, and returns a binary score and an explanation. The benchmark score is interpretable because every point corresponds to a concrete, inspectable decision.

\paragraph{Latent reaction points.}
\benchmark{} is a forecasting benchmark, not an entailment benchmark. We do not restrict criteria to reactions directly implied by the event description alone. We therefore include reaction points that real users surfaced through platform conventions, prior brand knowledge, local memes, demographic identity, and recent discourse.

% The target is \emph{endpoint reaction coverage}: given the event brief, can a model recover the high-salience reaction points that surfaced in public discourse? This measures marketing-relevant public attention rather than all privately possible reactions. Excluding latent anchors would turn the task back into controlled reasoning and miss the reactions that drive public attention. A criterion is valid when it corresponds to an observed reaction pattern, is abstracted away from raw user text, and can be applied consistently through its definition, examples, and judgment rule.

\begin{table}[htbp]
\centering
\caption{Example T1 flashpoint criterion for a product-recall topic. All 23{,}122 criteria follow this five-field structure.}
\label{tab:criterion_example}
\small
\begin{tabular}{@{}p{0.20\columnwidth}p{0.72\columnwidth}@{}}
\toprule
\rowcolor{CSBPanel}
\textbf{Field} & \textbf{Content} \\
\midrule
Trigger & ``Apology lacks sincerity'' \\
Definition & Users question whether the public apology is genuine or merely a tactic. \\
Positive example & ``This reads like their PR team wrote it'' \\
Negative example & ``I don't like this brand''  \\
Judgment rule & Credit only if the comment questions the \emph{sincerity or authenticity} of the apology, not general brand criticism \\
\bottomrule
\end{tabular}
\vspace{-1em}
\end{table}

\subsection{Dataset Collection and Construction}\label{sec:construction}
\paragraph{Data Collection.} We source data from RedNote (Xiaohongshu), a Chinese social platform with 300M+ monthly active users focused on lifestyle and consumer content, providing rich naturalistic context for consumer behavior. RedNote's trending-topic lists and topic summaries are published by multiple third-party public aggregators (e.g., \texttt{tophub.today}, \texttt{qian-gua.com}). We obtain hot-topic keywords and their public summaries from these aggregators and through manual collection. We only use publicly available content. We do not scrape or crawl RedNote, do not use any undocumented platform API, and do not access raw user posts, account identifiers, or any logged-in views. Each topic is verified by an actual public-trending search on the corresponding date, and the resulting artifact contains no quoted user text and no information by which an individual user could be identified. Collection complies with each aggregator's posted terms of service.

After this collection step we apply LLM-based filtering to retain only topics that are related to consumer reactions, and we manually sample 100 topics to verify the filtering quality.
\paragraph{Criteria drafting.} For each topic, an LLM proposes T1/T2/T3/T4 criteria with definitions, positive/negative examples, and judgment rules, along with objective topic descriptions. Each proposed criterion is anchored in observed source reactions, while an LLM standardizes those reactions into auditable records with explicit decision boundaries in Table \ref{tab:criterion_example}. Appendix~\ref{app:prompts} summarizes the operational drafting prompt.
\paragraph{Criteria hardening.} We harden the drafted criteria to remove event-restatement leakage, semantic duplication, over-pruning, and judge-fragile meme points. The process merges similar topics and criteria, drops criteria that collapse to the event description, checks within-topic independence, and audits pruned items. Appendix~\ref{app:hardening} gives the full hardening protocol and examples. After hardening and pruning, the released \textbf{evaluation set contains 23{,}122 criteria}. All headline leaderboard scores and error analyses in the paper are reported on this final evaluation set.

\section{Experiments and Results}
\subsection{Dataset Statistics}\label{sec:stats}
\begin{wraptable}{r}{0.38\textwidth}
\centering
\vspace{-4em}
\caption{Benchmark statistics.}
\vspace{0.5em}
\label{tab:dataset_stats}
\begin{tabular}{lr}
\toprule
\rowcolor{CSBPanel}
\textbf{Statistic} & \textbf{Value} \\
\midrule
Topic instances & 1,553 \\
Atomic criteria (total) & 23,122 \\
Sentiment flashpoints & 3,779 \\
Emotion keywords & 8,749 \\
Positive aspects & 5,666 \\
Negative aspects & 4,928 \\
Average criteria per topic & 14.9 \\
Median criteria per topic & 14 \\
\bottomrule
\end{tabular}
\vspace{-2em}
\end{wraptable}
We collected hot topics between March and September 2025. The final release contains 1,553 topics with 23,122 criteria. We release the canonical Chinese benchmark, English sidecars for reviewer inspection, judge prompts, and the evaluation harness under CC~BY-NC. Official evaluation uses the Chinese criteria for judge consistency, and the released criteria contain abstracted reaction-type descriptions rather than raw user content or account identifiers.
Additional statistics are shown in Table~\ref{tab:dataset_stats}.

\paragraph{Topic coverage.} The 1,553 topics span four categories and twenty consumer-facing sub-fields (Figure~\ref{fig:hero}, left). Society and public-issue topics dominate (37\%) because they generate the highest engagement on the platform, followed by entertainment (33\%), sports and gaming (17\%), and lifestyle goods (12\%). This breadth makes domain-specific strengths visible rather than letting any single topic family dominate the leaderboard.

\paragraph{Candidates.} We evaluate Gemini-3.1-Pro, Gemini-3-Flash, Qwen3.5-397B, Grok-4.2, DeepSeek-V3, Doubao-Seed-1.8-Thinking, Kimi-K2.5, GPT-5.2, MiniMax-M2.5, Claude-Opus-4.6, MiMo-V2.5-Pro, Claude-Sonnet-4.6, and GPT-4o.
\paragraph{Judge.} We use GPT-5.2 as the fixed judge for cost-effectiveness; \S\ref{sec:judge} validates judge reliability with cross-judge and human-audit checks.

\subsection{Main Results}\label{sec:main-results}
We now answer the opening question empirically. Can LLMs think like consumers, in the sense of reconstructing what a crowd will say in real public discussion? The headline is \textbf{not yet, and the remaining gap is not uniform}.

\begin{figure*}[t]
    \centering
    \vspace{-1em}\includegraphics[width=\textwidth]{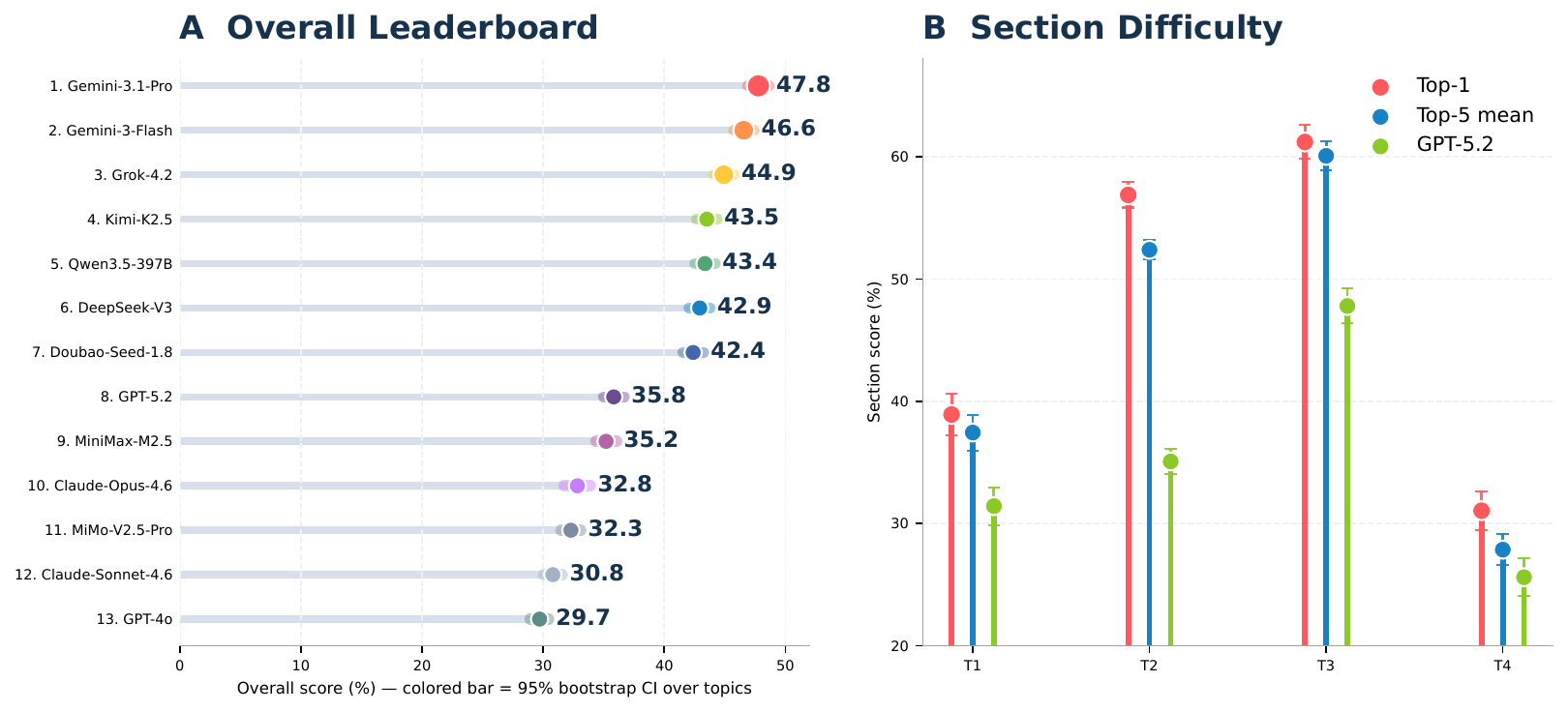}
    \vspace{-2em}
    \caption{Main results. Left: overall leaderboard on the final full benchmark. Right: section-wise difficulty, comparing the top model, the top-5 mean, and GPT-5.2. Flashpoints and negative aspects remain the hardest families.}
    \label{fig:leaderboard}
    \vspace{-1em}
\end{figure*}

Figure~\ref{fig:leaderboard} shows the full leaderboard. The best model, Gemini-3.1-Pro, reaches 47.8\% overall, followed by Gemini-3-Flash at 46.6\%. Grok-4.2, Kimi-K2.5, Qwen3.5-397B, DeepSeek-V3, and Doubao-Seed-1.8 form a tight 42--45\% middle tier. GPT-5.2 ranks eighth at 35.8\%; MiniMax-M2.5, Claude-Opus-4.6, MiMo-V2.5-Pro, Claude-Sonnet-4.6, and GPT-4o stay at 29--35\% in this setting. Rank order is stable under bootstrap resampling, and the broad tiering is stable under the criteria-weighted micro-average. The gap is not regional, and the hard sections (T1 flashpoints and T4 criticisms) remain difficult across the leaderboard.

Despite having competitive performance on technical benchmarks, top-tier models (such as Gemini-3.1-Pro, GPT-5.2 and Claude-Opus-4.6) \textbf{have a significant performance gap} in mimicking consumer behaviors. Gemini is roughly 12 points above GPT-5.2 and Claude-Opus-4.6, giving measurable form to a common user-facing complaint about emotional value, namely that some frontier assistants can produce fluent comments while still missing the affective targets and criticism vectors that make public reactions feel human. Moreover, \textbf{all models remain far from the ceiling}. Even the strongest model leaves nearly half of real consumer reactions uncovered. 

\begin{table}[t]
\centering
\caption{Full leaderboard on the final benchmark (1{,}553 topics, 23{,}122 criteria). Scores are percentages (judge GPT-5.2). Overall scores include 95\% bootstrap CIs over topics; the broad tier separation is stable under topic bootstrap and criteria-weighted micro-averaging.}
\vspace{-0.5em}
\label{tab:leaderboard}
\small
\begin{tabular}{rlccccc}
\toprule
\rowcolor{CSBPanel}
& \textbf{Model} & \textbf{\avg (95\% CI)} & \textbf{T1} & \textbf{T2} & \textbf{T3} & \textbf{T4} \\
\midrule
1 & Gemini-3.1-Pro & \textbf{47.8} {\small[46.9, 48.7]} & 38.9 & \textbf{56.9} & 61.2 & \textbf{31.1} \\
2 & Gemini-3-Flash & 46.6 {\small[45.7, 47.4]} & \textbf{39.5} & 53.1 & \textbf{63.8} & 26.8 \\
3 & Grok-4.2 & 44.9 {\small[44.1, 45.8]} & 36.8 & 52.5 & 60.0 & 27.2 \\
4 & Kimi-K2.5 & 43.5 {\small[42.7, 44.4]} & 35.9 & 49.0 & 55.6 & 30.4 \\
5 & Qwen3.5-397B & 43.4 {\small[42.5, 44.2]} & 36.1 & 50.4 & 59.8 & 23.9 \\
6 & DeepSeek-V3 & 42.9 {\small[42.0, 43.8]} & 33.9 & 49.2 & 54.8 & 30.6 \\
7 & Doubao-Seed-1.8 & 42.4 {\small[41.5, 43.2]} & 34.8 & 51.0 & 58.3 & 22.1 \\
\midrule
8 & GPT-5.2 & 35.8 {\small[35.0, 36.7]} & 31.4 & 35.1 & 47.8 & 25.6 \\
9 & MiniMax-M2.5 & 35.2 {\small[34.3, 36.1]} & 28.8 & 40.6 & 48.3 & 19.6 \\
10 & Claude-Opus-4.6 & 32.8 {\small[31.7, 33.9]} & 25.4 & 37.7 & 44.7 & 19.8 \\
11 & MiMo-V2.5-Pro & 32.3 {\small[31.5, 33.1]} & 27.4 & 39.7 & 45.3 & 20.1 \\
12 & Claude-Sonnet-4.6 & 30.8 {\small[30.0, 31.6]} & 24.8 & 32.6 & 46.0 & 16.1 \\
13 & GPT-4o & 29.7 {\small[28.9, 30.5]} & 23.5 & 35.8 & 43.5 & 12.1 \\
\bottomrule
\end{tabular}
\end{table}

\subsection{The Asymmetric Failure: Tone without Target}
\label{sec:asymmetric}

We find that \textbf{current LLMs can partially \emph{``sound''} like consumers, while failing to predict what consumers will focus on or attack.} Across all 13 models, T2 and T3 are 15--30 points easier than T1 and T4. For the top model, section scores have a 22--30 point spread that does not close at scale. It is a \emph{crowd-forecasting error}, where models often sound socially fluent, but fail to recover the concrete objects of attention, such as the meme phrases, product details, regional constraints, or business-model facts that become the public hook. For downstream marketing use cases, this is the decisive failure, because crisis forecasting and pre-launch risk assessment depend precisely on anticipating those hooks. The same T1/T4 gap persists across model tiers.
The same asymmetric pattern appears in a second-platform YouTube pilot (Appendix~\ref{app:youtube_pilot}, Figure~\ref{fig:youtube_pilot}), where the construction is instantiated on 100 English trending videos and again finds positive aspects easiest while flashpoints and negative aspects remain hardest.

\begin{wrapfigure}{l}{0.43\textwidth}
    \centering
    \includegraphics[width=\linewidth]{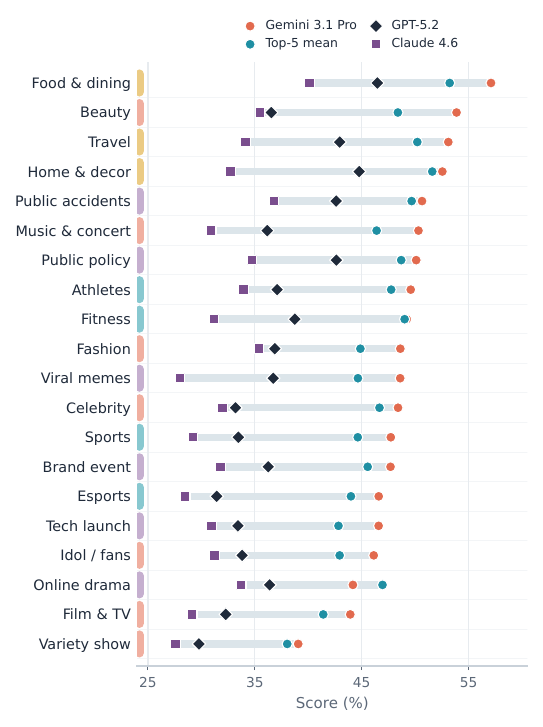}
    \vspace{-2em}
    \caption{Ranked sub-field gap plot across 20 consumer-facing domains. Rows are sorted by Gemini-3.1-Pro score; horizontal spans show the cross-model range.}
    \label{fig:subfield_heatmap}
    \vspace{1em}
\includegraphics[width=\linewidth]{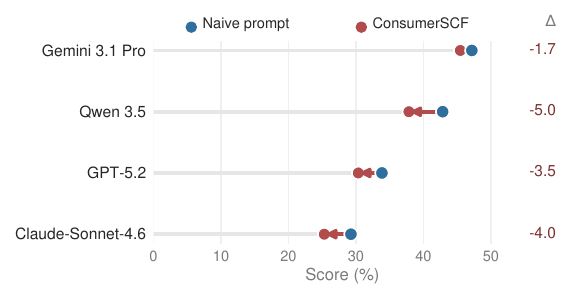}
    \caption{Direct SCF prompting does not improve coverage. Arrows show the change from naive prompting (blue) to \model{} (red).}
    \vspace{-3em}
    \label{fig:scf_ablation}
\end{wrapfigure}

Figure~\ref{fig:subfield_heatmap} further checks whether the failure is an artifact of one domain, and finds the gap is broad rather than domain-local. Model scores vary across the 20 sub-fields, but every sub-field remains far from ceiling, and the relative rankings are stable across domains. This suggests that the gap is not reducible to a single topic family such as celebrity gossip or tech launches.

\vspace{-1em}
\section{Ablations and Robustness}
\vspace{-0.5em}
\subsection{Where does the error come from?}\label{sec:errors}
We classify all 197{,}790 missed criteria across the 13 generators by parsing judge explanations into an 8-category taxonomy (Appendix~\ref{app:errors}; the categorization is derived from keyword patterns in judge text plus a 50-sample manual audit). Using this taxonomy, we find that \textbf{missing-content patterns dominate the error taxonomy}. Generations are topically relevant but do not surface the specific reaction trigger (44.5\% of misses), section-specific vocabulary at T2--T4 (28.6\%) or the flashpoint anchor itself (7.1\%). Combined, 80.2\% of misses pattern-match to lexical / anchor specificity rather than semantic disagreement. Meme repetition without pragmatic transfer (8.8\%) concentrates in T1; opposite-valence patterns (7.1\%) are most prevalent in T3. On the final 23{,}122-criterion benchmark, 6{,}677 criteria (28.9\%) are covered by zero of the 13 models. The two ``hard'' families have higher zero-coverage rates (T1 = 49.3\%, T4 = 41.6\%) than the easier two (T2 = 19.4\%, T3 = 18.8\%), consistent with flashpoint anticipation and criticism forecasting being the frontier challenges. These errors point to a more specific cognitive gap than generic sentiment modeling: models often recognize the event topic and produce plausible affect, but fail at \emph{attention-anchor selection}---deciding which concrete detail will become the object of public repetition, moralization, joking, or complaint. This also explains why T1 and T4 are hardest: they require stance-taking around a socially charged anchor, not merely naming an emotion or listing a product attribute.
\vspace{-2em}
\subsection{Does direct SCF prompting help?}
\label{sec:scaffold-negative}
We test whether a specially designed prompt can improve coverage. A standard ``few-shot'' control is not clean because the task has no canonical ground-truth response. There are no so-called ``reference comments''.

Instead, we develop \model{}, an expanded structured social-cognitive prompt based on consumer psychology. Before generation, it asks the model to internally reason through three social-theory lenses---self-presentation strategy, displayed capital, and framing~\citep{goffman1959presentation, bourdieu1984distinction, tversky1979prospect}---and operationalizes each lens with concrete participation motives, identity signals, resource cues, and narrative frames; the full prompt is shown in Appendix~\ref{scf}. This makes \model{} stronger than a vague role prompt, since it expands social cognition into participation motives, identity and resource signals, and narrative packaging choices. We randomly select 100 topics across four generators that span the leaderboard (Gemini-3.1-Pro, Qwen3.5-397B, GPT-5.2, Claude-Sonnet-4.6). As Figure~\ref{fig:scf_ablation} shows, \model{} averages \textbf{$-3.5$ pt} below the naive prompt, and \emph{all four generators} are worse (Gemini-3.1-Pro $-1.7$; Qwen3.5-397B $-5.0$; GPT-5.2 $-3.5$; Claude-Sonnet-4.6 $-4.0$). The result does not rule out every possible prompting variant, but it shows that simply inserting an explicit social-theory scaffold into a single generation call is not enough; under this direct formulation, coverage consistently falls.

We also evaluate a generate--reflect multi-agent pipeline based on \textsc{MegaAgent}~\citep{wang2025megaagent} on the same 100-topic slice. It first produces the naive comment set, then lets a CEO agent autonomously define subordinate agents, exchange files and messages, identify missing reaction angles, and refine the draft. It improves MiMo-V2.5-Pro from 32.9\% to \textbf{37.6\%} (+4.7 pt; 54 wins, 22 ties, 24 losses) and GPT-5.2 from 30.1\% to \textbf{31.8\%} (+1.8 pt; 41 wins, 29 ties, 30 losses). These gains show that \benchmark{} can reward iterative refinement beyond a single call.

Generic chain-of-thought is difficult to isolate for current frontier APIs, where hidden reasoning is already part of the serving stack. Text classifiers, ABSA, and opinion-mining models are natural baselines for fixed-label reaction prediction \citep{bojic2026llm}, but they predict labels for given text, whereas \benchmark{} requires open-ended comments for a new topic and then audits whether those comments recover unseen reaction criteria. 

\subsection{Are pointwise judges reliable?}\label{sec:judge}
Our main experiments use GPT-5.2 as the fixed judge for cost-effectiveness. We validate the scoring pipeline at three levels: inter-judge agreement, intra-judge stability, and human audit of pointwise labels. The audit asks annotators to assign their own yes/no label for each generated-comment/criterion pair, allowing us to compare frozen judge labels against independent human-majority labels.

For inter-judge reliability, we employ three independent judges (GPT-5.2, Gemini-3-Pro, and Claude-4.5-Opus) on 100+ criteria from 10 random topics with the same rules. The three judges achieve 92.1\% agreement. Compared with holistic LLM-as-Judge scoring, whose agreement rate collapses to 65.8\%, \textbf{our yes-no criteria yield substantially higher agreement}. We also show the necessity of our five-field criteria by keeping only the trigger itself: the agreement rate drops from 92.1\% to 84.8\%, showing that the judgment rule carries real signal. For intra-judge stability, we run GPT-5.2 three times on the same set and obtain 91.8\% agreement. Together, these checks support pointwise judging as a relatively stable method.

Finally, we connect inter-judge reliability to human validity by asking three human annotators to label the same 100+ criteria. The audit finds 98.4\% agreement between pointwise judge decisions and human-majority labels. On the hardest subset---the three-judge-disagreement cases---human majority still agrees with the LLM-judge majority in 84.6\% of cases. The remaining disagreements are semantic-boundary keyword cases, suggesting that errors are localized to ambiguous criterion boundaries rather than systematic judge misalignment. We also reduce length bias by judging only the first 3{,}600 characters of each generation and reduce style bias by asking the judge to apply criterion-specific definitions, examples, negative examples, and semantic-equivalence rules rather than rewarding rhetorical fluency. As a post-hoc check, verbosity alone does not explain the leaderboard: Gemini-3.1-Pro is the top model without being the longest generator, and MiMo-V2.5-Pro outperforms several longer generators.

\section{Release and Responsible Use}
\label{sec:release}
For responsible release, we use only publicly available topic-level materials and do not crawl RedNote posts, access undocumented platform APIs, or release raw user threads or account identifiers.

The public release at \url{https://huggingface.co/datasets/wty500/ConsumerSimBench} ships the benchmark JSONL, three judge prompts, English sidecars for reviewer inspection, and the evaluation harness. Its intended use is marketing-facing reaction forecasting: anticipating praise targets, criticism vectors, emotional flashpoints, and breakout angles before release. Raw RedNote threads are not included. A high score indicates reaction-forecasting capability, while deployment should still account for platform, product, and governance context (Appendix~\ref{app:limitations}).

\section{Conclusion}
\emph{Can LLMs think like consumers?} On \benchmark{}, the answer is \textbf{not yet}. The best system reconstructs only 47.8\% of what real public discussion surfaces, and the failure is asymmetric: models can sound emotional while failing to anticipate which trigger will dominate or which dimension users will attack. Moreover, direct structured prompting fails, while a multi-agent pipeline slightly improves. Closing the gap likely requires more than adding prompt-level scaffolding alone.

The broader message is a reframing, from \emph{dyadic belief inference} in fictional vignettes to \emph{crowd-level reaction reconstruction} in real public discourse. Models that sound consumer-like are not automatically models that anticipate consumer criticism, and for downstream marketing users, the latter is what matters. With top-tier LLMs surging on technical benchmarks, \benchmark{} gives measurable form to a complementary frontier of open-form humanities problems, where models must anticipate socially grounded reactions rather than solve formally specified tasks. This echoes users' complaint that today's models often lack real emotional value, not just logical competence.

\bibliographystyle{unsrt}   
\bibliography{references}

@article{mou2024individual,
  title={From individual to society: A survey on social simulation driven by large language model-based agents},
  author={Mou, Xinyi and Ding, Xuanwen and He, Qi and Wang, Liang and Liang, Jingcong and Zhang, Xinnong and Sun, Libo and Lin, Jiayu and Zhou, Jie and Huang, Xuanjing and others},
  journal={arXiv preprint arXiv:2412.03563},
  year={2024}
}

@article{hewitt2024predicting,
  title={Predicting results of social science experiments using large language models},
  author={Hewitt, Luke and Ashokkumar, Ashwini and Ghezae, Isaias and Willer, Robb},
  journal={Preprint},
  year={2024}
}

@inproceedings{tu2024charactereval,
  title={Charactereval: A chinese benchmark for role-playing conversational agent evaluation},
  author={Tu, Quan and Fan, Shilong and Tian, Zihang and Shen, Tianhao and Shang, Shuo and Gao, Xin and Yan, Rui},
  booktitle={Proceedings of the 62nd Annual Meeting of the Association for Computational Linguistics (Volume 1: Long Papers)},
  pages={11836--11850},
  year={2024}
}

@inproceedings{wang-etal-2024-rolellm,
    title = "{R}ole{LLM}: Benchmarking, Eliciting, and Enhancing Role-Playing Abilities of Large Language Models",
    author = "Wang, Noah  and
      Peng, Z.y.  and
      Que, Haoran  and
      Liu, Jiaheng  and
      Zhou, Wangchunshu  and
      Wu, Yuhan  and
      Guo, Hongcheng  and
      Gan, Ruitong  and
      Ni, Zehao  and
      Yang, Jian  and
      Zhang, Man  and
      Zhang, Zhaoxiang  and
      Ouyang, Wanli  and
      Xu, Ke  and
      Huang, Wenhao  and
      Fu, Jie  and
      Peng, Junran",
    editor = "Ku, Lun-Wei  and
      Martins, Andre  and
      Srikumar, Vivek",
    booktitle = "Findings of the Association for Computational Linguistics: ACL 2024",
    month = aug,
    year = "2024",
    address = "Bangkok, Thailand",
    publisher = "Association for Computational Linguistics",
    url = "https://aclanthology.org/2024.findings-acl.878/",
    doi = "10.18653/v1/2024.findings-acl.878",
    pages = "14743--14777"
}

@article{liu2024roleagent,
  title={Roleagent: Building, interacting, and benchmarking high-quality role-playing agents from scripts},
  author={Liu, Jiaheng and Ni, Zehao and Que, Haoran and Sun, Tao and Wang, Zekun and Yang, Jian and Wang, Jiakai and Guo, Hongcheng and Peng, Zhongyuan and Zhang, Ge and others},
  journal={Advances in Neural Information Processing Systems},
  volume={37},
  pages={49403--49428},
  year={2024}
}

@article{hu2025simbench,
  title={Simbench: Benchmarking the ability of large language models to simulate human behaviors},
  author={Hu, Tiancheng and Baumann, Joachim and Lupo, Lorenzo and Collier, Nigel and Hovy, Dirk and R{\"o}ttger, Paul},
  journal={arXiv preprint arXiv:2510.17516},
  year={2025}
}

@inproceedings{Park2023GenerativeAgents,
  author = {Park, Joon Sung and O'Brien, Joseph C. and Cai, Carrie J. and Morris, Meredith Ringel and Liang, Percy and Bernstein, Michael S.},
  title = {Generative Agents: Interactive Simulacra of Human Behavior},
  year = {2023},
  publisher = {Association for Computing Machinery},
  booktitle = {Proceedings of the 36th Annual ACM Symposium on User Interface Software and Technology},
  series = {UIST '23}
}

@article{gao2024large,
  title={Large language models empowered agent-based modeling and simulation: A survey and perspectives},
  author={Gao, Chen and Lan, Xiaochong and Li, Nian and Yuan, Yuan and Ding, Jingtao and Zhou, Zhilun and Xu, Fengli and Li, Yong},
  journal={Humanities and Social Sciences Communications},
  volume={11},
  number={1},
  pages={1--24},
  year={2024},
  publisher={Palgrave}
}

@inproceedings{chu2025llm,
  title={LLM-Based Multi-Agent System for Simulating and Analyzing Marketing and Consumer Behavior},
  author={Chu, Man-Lin and others},
  booktitle={IEEE International Conference on e-Business Engineering (ICEBE)},
  year={2025}
}

@article{anthis2025llm,
  title={Llm social simulations are a promising research method},
  author={Anthis, Jacy Reese and Liu, Ryan and Richardson, Sean M and Kozlowski, Austin C and Koch, Bernard and Evans, James and Brynjolfsson, Erik and Bernstein, Michael},
  journal={arXiv preprint arXiv:2504.02234},
  year={2025}
}

@article{hua2023war,
  title={War and peace (waragent): Large language model-based multi-agent simulation of world wars},
  author={Hua, Wenyue and Fan, Lizhou and Li, Lingyao and Mei, Kai and Ji, Jianchao and Ge, Yingqiang and Hemphill, Libby and Zhang, Yongfeng},
  journal={arXiv preprint arXiv:2311.17227},
  year={2023}
}

@article{wang2025user,
  title={User behavior simulation with large language model-based agents},
  author={Wang, Lei and Zhang, Jingsen and Yang, Hao and Chen, Zhi-Yuan and Tang, Jiakai and Zhang, Zeyu and Chen, Xu and Lin, Yankai and Sun, Hao and Song, Ruihua and others},
  journal={ACM Transactions on Information Systems},
  volume={43},
  number={2},
  pages={1--37},
  year={2025},
  publisher={ACM New York, NY}
}

@inproceedings{tang2025gensim,
  title={Gensim: A general social simulation platform with large language model based agents},
  author={Tang, Jiakai and Gao, Heyang and Pan, Xuchen and Wang, Lei and Tan, Haoran and Gao, Dawei and Chen, Yushuo and Chen, Xu and Lin, Yankai and Li, Yaliang and others},
  booktitle={Proceedings of the 2025 Conference of the Nations of the Americas Chapter of the Association for Computational Linguistics: Human Language Technologies (System Demonstrations)},
  pages={143--150},
  year={2025}
}

@article{zhang2025socioverse,
  title={Socioverse: A world model for social simulation powered by llm agents and a pool of 10 million real-world users},
  author={Zhang, Xinnong and Lin, Jiayu and Mou, Xinyi and Yang, Shiyue and Liu, Xiawei and Sun, Libo and Lyu, Hanjia and Yang, Yihang and Qi, Weihong and Chen, Yue and others},
  journal={arXiv preprint arXiv:2504.10157},
  year={2025}
}

@article{yang2024social,
  title={Social skill training with large language models},
  author={Yang, Diyi and Ziems, Caleb and Held, William and Shaikh, Omar and Bernstein, Michael S and Mitchell, John},
  journal={arXiv preprint arXiv:2404.04204},
  year={2024}
}

@inproceedings{wang2025megaagent,
  title={MegaAgent: A large-scale autonomous LLM-based multi-agent system without predefined SOPs},
  author={Wang, Qian and Wang, Tianyu and Tang, Zhenheng and Li, Qinbin and Chen, Nuo and Liang, Jingsheng and He, Bingsheng},
  booktitle={Findings of the Association for Computational Linguistics: ACL 2025},
  pages={4998--5036},
  year={2025}
}

@article{yang2025twinmarket,
  title={TwinMarket: A Scalable Behavioral and Social Simulation for Financial Markets},
  author={Yang, Yuzhe and Zhang, Yifei and Wu, Minghao and Zhang, Kaidi and Zhang, Yunmiao and Yu, Honghai and Hu, Yan and Wang, Benyou},
  journal={arXiv preprint arXiv:2502.01506},
  year={2025}
}

@article{piao2025agentsociety,
  title={Agentsociety: Large-scale simulation of llm-driven generative agents advances understanding of human behaviors and society},
  author={Piao, Jinghua and Yan, Yuwei and Zhang, Jun and Li, Nian and Yan, Junbo and Lan, Xiaochong and Lu, Zhihong and Zheng, Zhiheng and Wang, Jing Yi and Zhou, Di and others},
  journal={arXiv preprint arXiv:2502.08691},
  year={2025}
}

@inproceedings{li2024econagent,
  title={Econagent: large language model-empowered agents for simulating macroeconomic activities},
  author={Li, Nian and Gao, Chen and Li, Mingyu and Li, Yong and Liao, Qingmin},
  booktitle={Proceedings of the 62nd annual meeting of the association for computational linguistics (volume 1: Long papers)},
  pages={15523--15536},
  year={2024}
}

@article{gao2023s3,
  title={S3: Social-network simulation system with large language model-empowered agents},
  author={Gao, Chen and Lan, Xiaochong and Lu, Zhihong and Mao, Jinzhu and Piao, Jinghua and Wang, Huandong and Jin, Depeng and Li, Yong},
  journal={arXiv preprint arXiv:2307.14984},
  year={2023}
}

@inproceedings{zhang2025trendsim,
  title={Trendsim: Simulating trending topics in social media under poisoning attacks with llm-based multi-agent system},
  author={Zhang, Zeyu and Lian, Jianxun and Ma, Chen and Qu, Yaning and Luo, Ye and Wang, Lei and Li, Rui and Chen, Xu and Lin, Yankai and Wu, Le and others},
  booktitle={Findings of the Association for Computational Linguistics: NAACL 2025},
  pages={2930--2949},
  year={2025}
}

@article{yu2024fincon,
  title={Fincon: A synthesized llm multi-agent system with conceptual verbal reinforcement for enhanced financial decision making},
  author={Yu, Yangyang and Yao, Zhiyuan and Li, Haohang and Deng, Zhiyang and Jiang, Yuechen and Cao, Yupeng and Chen, Zhi and Suchow, Jordan and Cui, Zhenyu and Liu, Rong and others},
  journal={Advances in Neural Information Processing Systems},
  volume={37},
  pages={137010--137045},
  year={2024}
}

@article{yang2024oasis,
  title={Oasis: Open agent social interaction simulations with one million agents},
  author={Yang, Ziyi and Zhang, Zaibin and Zheng, Zirui and Jiang, Yuxian and Gan, Ziyue and Wang, Zhiyu and Ling, Zijian and Chen, Jinsong and Ma, Martz and Dong, Bowen and others},
  journal={arXiv preprint arXiv:2411.11581},
  year={2024}
}

@article{zhou2023sotopia,
  title={Sotopia: Interactive evaluation for social intelligence in language agents},
  author={Zhou, Xuhui and Zhu, Hao and Mathur, Leena and Zhang, Ruohong and Yu, Haofei and Qi, Zhengyang and Morency, Louis-Philippe and Bisk, Yonatan and Fried, Daniel and Neubig, Graham and others},
  journal={arXiv preprint arXiv:2310.11667},
  year={2023}
}

@article{wang2025customer,
  title={Customer-R1: Personalized simulation of human behaviors via RL-based LLM agent in online shopping},
  author={Wang, Ziyi and Lu, Yuxuan and Zhang, Yimeng and Huang, Jing and Wang, Dakuo},
  journal={arXiv preprint arXiv:2510.07230},
  year={2025}
}

@inproceedings{sap2019social,
  title={Social IQa: Commonsense Reasoning about Social Interactions},
  author={Sap, Maarten and Rashkin, Hannah and Chen, Derek and Le Bras, Ronan and Choi, Yejin},
  booktitle={Proceedings of the 2019 Conference on Empirical Methods in Natural Language Processing and the 9th International Joint Conference on Natural Language Processing (EMNLP-IJCNLP)},
  pages={4463--4473},
  year={2019}
}

@article{chuang2025debate,
  title={DEBATE: A Large-Scale Benchmark for Role-Playing LLM Agents in Multi-Agent, Long-Form Debates},
  author={Chuang, Yun-Shiuan and Tu, Ruixuan and Dai, Chengtao and Vasani, Smit and Yao, Binwei and Tessler, Michael Henry and Yang, Sijia and Shah, Dhavan and Hawkins, Robert and Hu, Junjie and others},
  journal={arXiv preprint arXiv:2510.25110},
  year={2025}
}

@inproceedings{
hendrycks2021aligning,
title={Aligning {\{}AI{\}} With Shared Human Values},
author={Dan Hendrycks and Collin Burns and Steven Basart and Andrew Critch and Jerry Li and Dawn Song and Jacob Steinhardt},
booktitle={International Conference on Learning Representations},
year={2021},
url={https://openreview.net/forum?id=dNy_RKzJacY}
}

@article{shao2023character,
  title={Character-llm: A trainable agent for role-playing},
  author={Shao, Yunfan and Li, Linyang and Dai, Junqi and Qiu, Xipeng},
  journal={arXiv preprint arXiv:2310.10158},
  year={2023}
}

@article{chenpersona,
  title={From Persona to Personalization: A Survey on Role-Playing Language Agents},
  author={Chen, Jiangjie and Wang, Xintao and Xu, Rui and Yuan, Siyu and Zhang, Yikai and Shi, Wei and Xie, Jian and Li, Shuang and Yang, Ruihan and Zhu, Tinghui and others},
  journal={Transactions on Machine Learning Research}
}

@inproceedings{wang2024incharacter,
  title={Incharacter: Evaluating personality fidelity in role-playing agents through psychological interviews},
  author={Wang, Xintao and Xiao, Yunze and Huang, Jen-tse and Yuan, Siyu and Xu, Rui and Guo, Haoran and Tu, Quan and Fei, Yaying and Leng, Ziang and Wang, Wei and others},
  booktitle={Proceedings of the 62nd annual meeting of the association for computational linguistics (volume 1: Long papers)},
  pages={1840--1873},
  year={2024}
}

@inproceedings{samuel-etal-2025-personagym,
    title = "{P}ersona{G}ym: Evaluating Persona Agents and {LLM}s",
    author = "Samuel, Vinay  and
      Zou, Henry Peng  and
      Zhou, Yue  and
      Chaudhari, Shreyas  and
      Kalyan, Ashwin  and
      Rajpurohit, Tanmay  and
      Deshpande, Ameet  and
      Narasimhan, Karthik R  and
      Murahari, Vishvak",
    editor = "Christodoulopoulos, Christos  and
      Chakraborty, Tanmoy  and
      Rose, Carolyn  and
      Peng, Violet",
    booktitle = "Findings of the Association for Computational Linguistics: EMNLP 2025",
    month = nov,
    year = "2025",
    address = "Suzhou, China",
    publisher = "Association for Computational Linguistics",
    url = "https://aclanthology.org/2025.findings-emnlp.368/",
    doi = "10.18653/v1/2025.findings-emnlp.368",
    pages = "6999--7022",
    ISBN = "979-8-89176-335-7"
}

@inproceedings{lee2025llms,
  title={Do llms have distinct and consistent personality? trait: Personality testset designed for llms with psychometrics},
  author={Lee, Seungbeen and Lim, Seungwon and Han, Seungju and Oh, Giyeong and Chae, Hyungjoo and Chung, Jiwan and Kim, Minju and Kwak, Beong-woo and Lee, Yeonsoo and Lee, Dongha and others},
  booktitle={Findings of the Association for Computational Linguistics: NAACL 2025},
  pages={8397--8437},
  year={2025}
}

@article{zeng2025psychcounsel,
  title={PsychCounsel-Bench: Evaluating the Psychology Intelligence of Large Language Models},
  author={Zeng, Min},
  journal={arXiv preprint arXiv:2510.01611},
  year={2025}
}

@inproceedings{shinoda2025tomato,
  title={Tomato: Verbalizing the mental states of role-playing llms for benchmarking theory of mind},
  author={Shinoda, Kazutoshi and Hojo, Nobukatsu and Nishida, Kyosuke and Mizuno, Saki and Suzuki, Keita and Masumura, Ryo and Sugiyama, Hiroaki and Saito, Kuniko},
  booktitle={Proceedings of the AAAI Conference on Artificial Intelligence},
  volume={39},
  number={2},
  pages={1520--1528},
  year={2025}
}

@inproceedings{emelin2021moral,
  title={Moral stories: Situated reasoning about norms, intents, actions, and their consequences},
  author={Emelin, Denis and Le Bras, Ronan and Hwang, Jena D and Forbes, Maxwell and Choi, Yejin},
  booktitle={Proceedings of the 2021 Conference on Empirical Methods in Natural Language Processing},
  pages={698--718},
  year={2021}
}

@inproceedings{wangcustomer,
  title={Customer-R1: personalized simulation of Human Behaviors via RL-based LLM Agent in Online Shopping},
  author={Wang, Ziyi and Lu, Yuxuan and Zhang, Yimeng and Huang, Jing and Wang, Dakuo},
  booktitle={First Workshop on Multi-Turn Interactions in Large Language Models}
}

@inproceedings{chen2024socialbench,
  title={Socialbench: Sociality evaluation of role-playing conversational agents},
  author={Chen, Hongzhan and Chen, Hehong and Yan, Ming and Xu, Wenshen and Xing, Gao and Shen, Weizhou and Quan, Xiaojun and Li, Chenliang and Zhang, Ji and Huang, Fei},
  booktitle={Findings of the Association for Computational Linguistics: ACL 2024},
  pages={2108--2126},
  year={2024}
}

@inproceedings{wu2023hi,
  title={Hi-tom: A benchmark for evaluating higher-order theory of mind reasoning in large language models},
  author={Wu, Yufan and He, Yinghui and Jia, Yilin and Mihalcea, Rada and Chen, Yulong and Deng, Naihao},
  booktitle={Findings of the Association for Computational Linguistics: EMNLP 2023},
  pages={10691--10706},
  year={2023}
}

@inproceedings{kim2023fantom,
  title={FANToM: A benchmark for stress-testing machine theory of mind in interactions},
  author={Kim, Hyunwoo and Sclar, Melanie and Zhou, Xuhui and Bras, Ronan and Kim, Gunhee and Choi, Yejin and Sap, Maarten},
  booktitle={Proceedings of the 2023 Conference on Empirical Methods in Natural Language Processing},
  pages={14397--14413},
  year={2023}
}

@inproceedings{xu2024opentom,
  title={OpenToM: A Comprehensive Benchmark for Evaluating Theory-of-Mind Reasoning Capabilities of Large Language Models},
  author={Xu, Hainiu and Zhao, Runcong and Zhu, Lixing and Du, Jinhua and He, Yulan},
  booktitle={Proceedings of the 62nd Annual Meeting of the Association for Computational Linguistics (Volume 1: Long Papers)},
  pages={8593--8623},
  year={2024}
}

@article{strachan2024testing,
  title={Testing theory of mind in large language models and humans},
  author={Strachan, James WA and Albergo, Dalila and Borghini, Giulia and Pansardi, Oriana and Scaliti, Eugenio and Gupta, Saurabh and Saxena, Krati and Rufo, Alessandro and Panzeri, Stefano and Manzi, Guido and others},
  journal={Nature Human Behaviour},
  volume={8},
  number={7},
  pages={1285--1295},
  year={2024},
  publisher={Nature Publishing Group UK London}
}

@inproceedings{shi2025muma,
  title={Muma-tom: Multi-modal multi-agent theory of mind},
  author={Shi, Haojun and Ye, Suyu and Fang, Xinyu and Jin, Chuanyang and Isik, Leyla and Kuo, Yen-Ling and Shu, Tianmin},
  booktitle={Proceedings of the AAAI Conference on Artificial Intelligence},
  volume={39},
  number={2},
  pages={1510--1519},
  year={2025}
}

@inproceedings{chen2024emotionqueen,
  title={EmotionQueen: A Benchmark for Evaluating Empathy of Large Language Models},
  author={Chen, Yuyan and Yan, Songzhou and Liu, Sijia and Li, Yueze and Xiao, Yanghua},
  booktitle={Findings of the Association for Computational Linguistics ACL 2024},
  pages={2149--2176},
  year={2024}
}

@inproceedings{chen2025theory,
    title = "Theory of Mind in Large Language Models: Assessment and Enhancement",
    author = "Chen, Ruirui  and
      Jiang, Weifeng  and
      Qin, Chengwei  and
      Tan, Cheston",
    editor = "Che, Wanxiang  and
      Nabende, Joyce  and
      Shutova, Ekaterina  and
      Pilehvar, Mohammad Taher",
    booktitle = "Proceedings of the 63rd Annual Meeting of the Association for Computational Linguistics (Volume 1: Long Papers)",
    month = jul,
    year = "2025",
    address = "Vienna, Austria",
    publisher = "Association for Computational Linguistics",
    url = "https://aclanthology.org/2025.acl-long.1522/",
    doi = "10.18653/v1/2025.acl-long.1522",
    pages = "31539--31558",
    ISBN = "979-8-89176-251-0"
}

@article{liu2025mind,
  title={The mind in the machine: A survey of incorporating psychological theories in llms},
  author={Liu, Zizhou and Gong, Ziwei and Ai, Lin and Hui, Zheng and Chen, Run and Leach, Colin Wayne and Greene, Michelle R and Hirschberg, Julia},
  journal={arXiv preprint arXiv:2505.00003},
  year={2025}
}

@inproceedings{liu2024interintent,
  title={InterIntent: Investigating Social Intelligence of LLMs via Intention Understanding in an Interactive Game Context},
  author={Liu, Ziyi and Anand, Abhishek and Zhou, Pei and Huang, Jen-tse and Zhao, Jieyu},
  booktitle={Proceedings of the 2024 Conference on Empirical Methods in Natural Language Processing},
  pages={6718--6746},
  year={2024}
}

@article{gandhi2023understanding,
  title={Understanding social reasoning in language models with language models},
  author={Gandhi, Kanishk and Fr{\"a}nken, Jan-Philipp and Gerstenberg, Tobias and Goodman, Noah},
  journal={Advances in Neural Information Processing Systems},
  volume={36},
  pages={13518--13529},
  year={2023}
}

@article{zhang2024affective,
  title={Affective computing in the era of large language models: A survey from the nlp perspective},
  author={Zhang, Yiqun and Yang, Xiaocui and Xu, Xingle and Gao, Zeran and Huang, Yijie and Mu, Shiyi and Feng, Shi and Wang, Daling and Zhang, Yifei and Song, Kaisong and others},
  journal={arXiv preprint arXiv:2408.04638},
  year={2024}
}

@inproceedings{echterhoff2024cognitive,
  title={Cognitive bias in decision-making with LLMs},
  author={Echterhoff, Jessica Maria and Liu, Yao and Alessa, Abeer and McAuley, Julian and He, Zexue},
  booktitle={Findings of the association for computational linguistics: EMNLP 2024},
  pages={12640--12653},
  year={2024}
}

@inproceedings{chen2024agentverse,
  title={AgentVerse: Facilitating Multi-Agent Collaboration and Exploring Emergent Behaviors.},
  author={Chen, Weize and Su, Yusheng and Zuo, Jingwei and Yang, Cheng and Yuan, Chenfei and Chan, Chi-Min and Yu, Heyang and Lu, Yaxi and Hung, Yi-Hsin and Qian, Chen and others},
  booktitle={ICLR},
  year={2024}
}

@article{jia2024decision,
  title={Decision-making behavior evaluation framework for llms under uncertain context},
  author={Jia, Jingru and Yuan, Zehua and Pan, Junhao and McNamara, Paul E and Chen, Deming},
  journal={Advances in neural information processing systems},
  volume={37},
  pages={113360--113382},
  year={2024}
}

@article{jimenez2023swe,
  title={Swe-bench: Can language models resolve real-world github issues?},
  author={Jimenez, Carlos E and Yang, John and Wettig, Alexander and Yao, Shunyu and Pei, Kexin and Press, Ofir and Narasimhan, Karthik},
  journal={arXiv preprint arXiv:2310.06770},
  year={2023}
}

@article{merrill2026terminal,
  title={Terminal-bench: Benchmarking agents on hard, realistic tasks in command line interfaces},
  author={Merrill, Mike A and Shaw, Alexander G and Carlini, Nicholas and Li, Boxuan and Raj, Harsh and Bercovich, Ivan and Shi, Lin and Shin, Jeong Yeon and Walshe, Thomas and Buchanan, E Kelly and others},
  journal={arXiv preprint arXiv:2601.11868},
  year={2026}
}

@article{Klinkert_Buongiorno_Clark_2024, title={Evaluating the Efficacy of LLMs to Emulate Realistic Human Personalities}, volume={20}, url={https://ojs.aaai.org/index.php/AIIDE/article/view/31867}, DOI={10.1609/aiide.v20i1.31867}, abstractNote={To enhance immersion and engagement in video games, the design of Affective Non-Player Characters (NPCs) is a key focus for researchers and practitioners. Affective Computing frameworks improve Non-player characters (NPC) by providing personalities, emotions, and social relations. Large Language Models (LLMs) bring the promise to dynamically enhance character design when coupled with these frameworks, but further research is needed to validate the models truly represent human qualities. In this research, a comprehensive analysis investigates the capabilities of LLMs to generate content that aligns with human personality, using the Big Five and human responses from the International Personality Item Pool (IPIP) questionnaire. Our goal is to benchmark the performance of various LLMs, including frontier models and local models, against an extensive dataset comprising over 50,000 human surveys of self-reported personality tests to determine whether LLMs can replicate human-like decision-making with personality-driven prompts. A range of personality profiles were used to cluster the test results from the human survey dataset. Our methodology involved prompting LLMs with self-evaluated test items for each personality profile, comparing their outputs to human baseline responses, and evaluating the accuracy and consistency. Our findings show that some local models had 0% alignment of any personality profiles when compared to the human dataset, while the frontier models, in some cases, had 100% alignment. The results indicate that NPCs can successfully emulate human-like personality traits using LLMs, as demonstrated by benchmarking the LLM’s output against human data. This foundational work serves as a methodology for game developers and researchers to test and evaluate LLMs, ensuring they accurately represent the desired human personalities and can be expanded for further validation.}, number={1}, journal={Proceedings of the AAAI Conference on Artificial Intelligence and Interactive Digital Entertainment}, author={Klinkert, Lawrence J. and Buongiorno, Steph and Clark, Corey}, year={2024}, month={Nov.}, pages={65-75} }

@article{wang2024mmlu,
  title={Mmlu-pro: A more robust and challenging multi-task language understanding benchmark},
  author={Wang, Yubo and Ma, Xueguang and Zhang, Ge and Ni, Yuansheng and Chandra, Abhranil and Guo, Shiguang and Ren, Weiming and Arulraj, Aaran and He, Xuan and Jiang, Ziyan and others},
  journal={Advances in Neural Information Processing Systems},
  volume={37},
  pages={95266--95290},
  year={2024}
}

@article{zakazov2024assessing,
  title={Assessing Social Alignment: Do Personality-Prompted Large Language Models Behave Like Humans?},
  author={Zakazov, Ivan and Boronski, Mikolaj and Drudi, Lorenzo and West, Robert},
  journal={arXiv preprint arXiv:2412.16772},
  year={2024}
}

@inproceedings{marraffini2024greatest,
  title={The greatest good benchmark: Measuring llms’ alignment with utilitarian moral dilemmas},
  author={Marraffini, Giovanni Franco Gabriel and Cotton, Andr{\'e}s and Hsueh, Noe Fabian and Fridman, Axel and Wisznia, Juan and Del Corro, Luciano},
  booktitle={Proceedings of the 2024 Conference on Empirical Methods in Natural Language Processing},
  pages={21950--21959},
  year={2024}
}

@article{xiao2025evaluating,
  title={Evaluating the ability of large Language models to predict human social decisions},
  author={Xiao, Feng and Wang, XT XiaoTian},
  journal={Scientific Reports},
  volume={15},
  number={1},
  pages={32290},
  year={2025},
  publisher={Nature Publishing Group UK London}
}

@inproceedings{
anonymous2025socialr,
title={Social-R1: Enhancing Social Intelligence in {LLM}s through Human-like Reinforced Reasoning},
author={Anonymous},
booktitle={Submitted to The Fourteenth International Conference on Learning Representations},
year={2025},
url={https://openreview.net/forum?id=3qAzQyOOnA},
note={under review}
}

@article{zhou2025think,
  title={Think Socially via Cognitive Reasoning},
  author={Zhou, Jinfeng and Chen, Zheyu and Wang, Shuai and Dai, Quanyu and Dong, Zhenhua and Wang, Hongning and Huang, Minlie},
  journal={arXiv preprint arXiv:2509.22546},
  year={2025}
}

@article{hwang2025infusing,
  title={Infusing Theory of Mind into Socially Intelligent LLM Agents},
  author={Hwang, EunJeong and Yin, Yuwei and Carenini, Giuseppe and West, Peter and Shwartz, Vered},
  journal={arXiv preprint arXiv:2509.22887},
  year={2025}
}

@article{ong2025human,
  title={Human behavior atlas: Benchmarking unified psychological and social behavior understanding},
  author={Ong, Keane and Dai, Wei and Li, Carol and Feng, Dewei and Li, Hengzhi and Wu, Jingyao and Cheong, Jiaee and Mao, Rui and Mengaldo, Gianmarco and Cambria, Erik and others},
  journal={arXiv preprint arXiv:2510.04899},
  year={2025}
}

@article{yong2025motivebench,
  title={MotiveBench: How Far Are We From Human-Like Motivational Reasoning in Large Language Models?},
  author={Yong, Xixian and Lian, Jianxun and Yi, Xiaoyuan and Zhou, Xiao and Xie, Xing},
  journal={arXiv preprint arXiv:2506.13065},
  year={2025}
}

@inproceedings{sabour2024emobench,
  title={Emobench: Evaluating the emotional intelligence of large language models},
  author={Sabour, Sahand and Liu, Siyang and Zhang, Zheyuan and Liu, June and Zhou, Jinfeng and Sunaryo, Alvionna and Lee, Tatia and Mihalcea, Rada and Huang, Minlie},
  booktitle={Proceedings of the 62nd Annual Meeting of the Association for Computational Linguistics (Volume 1: Long Papers)},
  pages={5986--6004},
  year={2024}
}

@article{kang2025hssbench,
  title={HSSBench: Benchmarking Humanities and Social Sciences Ability for Multimodal Large Language Models},
  author={Kang, Zhaolu and Gong, Junhao and Yan, Jiaxu and Xia, Wanke and Wang, Yian and Wang, Ziwen and Ding, Huaxuan and Cheng, Zhuo and Cao, Wenhao and Feng, Zhiyuan and others},
  journal={arXiv preprint arXiv:2506.03922},
  year={2025}
}

@inproceedings{mou-etal-2025-agentsense,
    title = "{A}gent{S}ense: Benchmarking Social Intelligence of Language Agents through Interactive Scenarios",
    author = "Mou, Xinyi  and
      Liang, Jingcong  and
      Lin, Jiayu  and
      Zhang, Xinnong  and
      Liu, Xiawei  and
      Yang, Shiyue  and
      Ye, Rong  and
      Chen, Lei  and
      Kuang, Haoyu  and
      Huang, Xuanjing  and
      Wei, Zhongyu",
    editor = "Chiruzzo, Luis  and
      Ritter, Alan  and
      Wang, Lu",
    booktitle = "Proceedings of the 2025 Conference of the Nations of the Americas Chapter of the Association for Computational Linguistics: Human Language Technologies (Volume 1: Long Papers)",
    month = apr,
    year = "2025",
    address = "Albuquerque, New Mexico",
    publisher = "Association for Computational Linguistics",
    url = "https://aclanthology.org/2025.naacl-long.257/",
    doi = "10.18653/v1/2025.naacl-long.257",
    pages = "4975--5001",
    ISBN = "979-8-89176-189-6"
}

@article{gu2024llmasajudge,
   title     = {A Survey on {LLM-as-a-Judge}},
   author    = {Gu, Jiawei and Jiang, Xuhui and Shi, Zhichao and Tan, Hexiang and Zhai, Xuehao and Xu, Chengjin and Li, Wei and Shen, Yinghan and Ma, Shengjie and Liu, Honghao and Wang, Saizhuo and Zhang, Kun and Wang, Yuanzhuo and Gao, Wen and Ni, Lionel and Guo, Jian},
  journal   = {arXiv preprint arXiv:2411.15594},
   year      = {2024}
 }

@inproceedings{chen-etal-2024-humans,
  title     = {Humans or {LLM}s as the Judge? A Study on Judgement Bias},
  author    = {Chen, Guiming Hardy and Chen, Shunian and Liu, Ziche and Jiang, Feng and Wang, Benyou},
  booktitle = {Proceedings of the 2024 Conference on Empirical Methods in Natural Language Processing},
  pages     = {8301--8327},
  year      = {2024},
  publisher = {Association for Computational Linguistics}
}

@article{argyle2023out,
  title={Out of One, Many: Using Language Models to Simulate Human Samples},
  author={Argyle, Lisa P. and Busby, Ethan C. and Fulda, Nancy and Gubler, Joshua R. and Rytting, Christopher and Wingate, David},
  journal={Political Analysis},
  volume={31},
  number={3},
  pages={337--351},
  year={2023},
  publisher={Cambridge University Press}
}

@book{goffman1959presentation,
  title={The Presentation of Self in Everyday Life},
  author={Goffman, Erving},
  year={1959},
  publisher={Doubleday}
}

@book{bourdieu1984distinction,
  title={Distinction: A Social Critique of the Judgement of Taste},
  author={Bourdieu, Pierre},
  year={1984},
  publisher={Harvard University Press}
}

@article{tversky1979prospect,
  title={Prospect Theory: An Analysis of Decision under Risk},
  author={Kahneman, Daniel and Tversky, Amos},
  journal={Econometrica},
  volume={47},
  number={2},
  pages={263--291},
  year={1979}
}

@inproceedings{SMP2023,
  title={SMP Challenge: An Overview and Analysis of Social Media Prediction Challenge},
  author={Wu, Bo and Liu, Peiye and Cheng, Wen-Huang and Liu, Bei and Zeng, Zhaoyang and Wang, Jia and Huang, Qiushi and Luo, Jiebo},
  booktitle={Proceedings of the 31st ACM International Conference on Multimedia},
  year={2023}}

@article{maier2025llms,
  title={LLMs Reproduce Human Purchase Intent via Semantic Similarity Elicitation of Likert Ratings},
  author={Maier, Benjamin F and Aslak, Ulf and Fiaschi, Luca and Rismal, Nina and Fletcher, Kemble and Luhmann, Christian C and Dow, Robbie and Pappas, Kli and Wiecki, Thomas V},
  journal={arXiv preprint arXiv:2510.08338},
  year={2025}
}

@inproceedings{10.1145/3701716.3715258,
author = {Chen, Luyu and Dai, Quanyu and Zhang, Zeyu and Feng, Xueyang and Zhang, Mingyu and Tang, Pengcheng and Chen, Xu and Zhu, Yue and Dong, Zhenhua},
title = {RecUserSim: A Realistic and Diverse User Simulator for Evaluating Conversational Recommender Systems},
year = {2025},
isbn = {9798400713316},
publisher = {Association for Computing Machinery},
address = {New York, NY, USA},
url = {https://doi.org/10.1145/3701716.3715258},
doi = {10.1145/3701716.3715258},
booktitle = {Companion Proceedings of the ACM on Web Conference 2025},
pages = {133–142},
numpages = {10},
location = {Sydney NSW, Australia},
series = {WWW '25}
}

@article{lin2026large,
  title={Large language models as psychological simulators: A methodological guide},
  author={Lin, Zhicheng},
  journal={Advances in Methods and Practices in Psychological Science},
  volume={9},
  number={1},
  pages={25152459251410153},
  year={2026},
  publisher={SAGE Publications Sage CA: Los Angeles, CA}
}

@InProceedings{pmlr-v202-aher23a,
  title = {Using Large Language Models to Simulate Multiple Humans and Replicate Human Subject Studies},
  author = {Aher, Gati V and Arriaga, Rosa I. and Kalai, Adam Tauman},
  booktitle = {Proceedings of the 40th International Conference on Machine Learning},
  pages = {337--371},
  year = {2023},
  editor = {Krause, Andreas and Brunskill, Emma and Cho, Kyunghyun and Engelhardt, Barbara and Sabato, Sivan and Scarlett, Jonathan},
  volume = {202},
  series = {Proceedings of Machine Learning Research},
  month = {23--29 Jul},
  publisher = {PMLR},
  url = {https://proceedings.mlr.press/v202/aher23a.html}
}

@article{brand2023using,
  title={Using LLMs for market research},
  author={Brand, James and Israeli, Ayelet and Ngwe, Donald},
  journal={Harvard Business School Marketing Unit Working Paper},
  number={23-062},
  year={2023}
}

@article{bisbee2024synthetic,
  title={Synthetic replacements for human survey data? The perils of large language models},
  author={Bisbee, James and Clinton, Joshua D and Dorff, Cassy and Kenkel, Brenton and Larson, Jennifer M},
  journal={Political Analysis},
  volume={32},
  number={4},
  pages={401--416},
  year={2024},
  publisher={Cambridge University Press}
}

@article{binz2025foundation,
  title={A foundation model to predict and capture human cognition},
  author={Binz, Marcel and Akata, Elif and Bethge, Matthias and Br{\"a}ndle, Franziska and Callaway, Fred and Coda-Forno, Julian and Dayan, Peter and Demircan, Can and Eckstein, Maria K and {\'E}ltet{\H{o}}, No{\'e}mi and others},
  journal={Nature},
  volume={644},
  number={8078},
  pages={1002--1009},
  year={2025},
  publisher={Nature Publishing Group UK London}
}

@article{slumbers2025using,
  title={Using large language models to simulate human behavioural experiments: Port of mars},
  author={Slumbers, Oliver and Leibo, Joel Z and Janssen, Marco A},
  journal={arXiv preprint arXiv:2506.05555},
  year={2025}
}

@article{agarwal2025silicon,
  title={The Silicon Sample: Benchmarking Synthetic Users Against Human Respondents in Market Research},
  author={Agarwal, Arya},
  journal={Available at SSRN 5835122},
  year={2025}
}

@article{seshadri2026lost,
  title={Lost in Simulation: LLM-Simulated Users are Unreliable Proxies for Human Users in Agentic Evaluations},
  author={Seshadri, Preethi and Cahyawijaya, Samuel and Odumakinde, Ayomide and Singh, Sameer and Goldfarb-Tarrant, Seraphina},
  journal={arXiv preprint arXiv:2601.17087},
  year={2026}
}

@article{poole2025benchmarking,
  title={Benchmarking Overton Pluralism in LLMs},
  author={Poole-Dayan, Elinor and Wu, Jiayi and Sorensen, Taylor and Pei, Jiaxin and Bakker, Michiel A},
  journal={arXiv preprint arXiv:2512.01351},
  year={2025}
}

@article{qi2025cross,
  title={A Cross-Cultural Comparison of LLM-based Public Opinion Simulation: Evaluating Chinese and US Models on Diverse Societies},
  author={Qi, Weihong and Huang, Fan and An, Jisun and Kwak, Haewoon},
  journal={arXiv preprint arXiv:2506.21587},
  year={2025}
}

@article{ludwig2026extracting,
  title={Extracting Consumer Insight from Text: A Large Language Model Approach to Emotion and Evaluation Measurement},
  author={Ludwig, Stephan and Danaher, Peter J and Yang, Xiaohao and Lin, Yu-Ting and Abedin, Ehsan and Grewal, Dhruv and Du, Lan},
  journal={arXiv preprint arXiv:2602.15312},
  year={2026}
}

@article{wang2026large,
  title={Large language models for market research: A data-augmentation approach},
  author={Wang, Mengxin and Zhang, Dennis J and Zhang, Heng},
  journal={Marketing Science},
  year={2026},
  publisher={INFORMS}
}

@inproceedings{dou2025simulatorarena,
  title={SimulatorArena: Are User Simulators Reliable Proxies for Multi-Turn Evaluation of AI Assistants?},
  author={Dou, Yao and Galley, Michel and Peng, Baolin and Kedzie, Chris and Cai, Weixin and Ritter, Alan and Quirk, Chris and Xu, Wei and Gao, Jianfeng},
  booktitle={Proceedings of the 2025 Conference on Empirical Methods in Natural Language Processing},
  pages={35200--35278},
  year={2025}
}

@article{su2026sell,
  title={Sell More, Play Less: Benchmarking LLM Realistic Selling Skill},
  author={Su, Xuanbo and Hu, Wenhao and Zhan, Le and Yang, Yanqi and Huang, Leo},
  journal={arXiv preprint arXiv:2604.07054},
  year={2026}
}

@article{qu2024performance,
  title={Performance and Biases of Large Language Models in Public Opinion Simulation},
  author={Qu, Yao and Wang, Jue},
  journal={Humanities and Social Sciences Communications},
  volume={11},
  number={1},
  pages={1--13},
  year={2024},
  publisher={Palgrave}
}

@article{miranda2025simulating,
  title={Simulating Public Opinion: Comparing Distributional and Individual-Level Predictions from LLMs and Random Forests},
  author={Miranda, Fernando and Balbi, Pedro Paulo},
  journal={Entropy},
  volume={27},
  number={9},
  pages={923},
  year={2025},
  publisher={MDPI}
}

@article{liu2025can,
  title={Can AI Automatically Analyze Public Opinion? A LLM Agents-Based Agentic Pipeline for Timely Public Opinion Analysis},
  author={Liu, Jing and Ren, Xinxing and Xu, Yanmeng and Guo, Zekun},
  journal={arXiv preprint arXiv:2505.11401},
  year={2025}
}

@article{bojic2026llm,
  title={LLM Agents Predict Social Media Reactions but Do Not Outperform Text Classifiers: Benchmarking Simulation Accuracy Using 120K+ Personas of 1511 Humans},
  author={Bojic, Ljubisa and Felfernig, Alexander and Dinic, Bojana and Ilic, Velibor and Rettinger, Achim and Mevorah, Vera and Trilling, Damian},
  journal={arXiv preprint arXiv:2604.19787},
  year={2026}
}

@inproceedings{schwager2026towards,
  title={Towards Simulating Social Media Users with LLMs: Evaluating the Operational Validity of Conditioned Comment Prediction},
  author={Schwager, Nils and M{\"u}nker, Simon and Plum, Alistair and Rettinger, Achim},
  booktitle={The Proceedings for the 15th Workshop on Computational Approaches to Subjectivity, Sentiment Social Media Analysis (WASSA 2026)},
  pages={208--221},
  year={2026}
}

@article{wang2025llm,
  title={LLM-based Human Simulations Have Not Yet Been Reliable},
  author={Wang, Qian and Wu, Jiaying and Jiang, Zichen and Tang, Zhenheng and Luo, Bingqiao and Chen, Nuo and Chen, Wei and He, Bingsheng},
  journal={arXiv preprint arXiv:2501.08579},
  year={2025}
}

%%%%%%%%%%%%%%%%%%%%%%%%%%%%%%%%%%%%%%%%%%%%%%%%%%%%%%%%%%%%
\newpage
\appendix

\section{Criteria-Weighted Micro-Average}
\label{app:micro}
Table~\ref{tab:micro} reports the criteria-weighted micro-average as a secondary metric. For each model, the micro value is $\frac{1}{4}\sum_{s\in S} \frac{\sum_i \text{covered}_{i,s}}{\sum_i \text{total}_{i,s}}$, i.e., each section is averaged by total criterion count rather than topic count. Deviations from the balanced (macro) average reported in Table~\ref{tab:leaderboard} are at most 1.0 point for all 13 models, and the broad leaderboard tiers are unchanged. The paper's conclusions are therefore insensitive to this metric choice.

\begin{table}[h]
\centering
\caption{Macro (balanced, 25\% per family) vs.\ criteria-weighted micro average. Differences are ${\le}1.0$ point for all models; broad leaderboard tiers are preserved.}
\label{tab:micro}
\small
\begin{tabular}{lccc}
\toprule
\rowcolor{CSBPanel}
\textbf{Model} & \textbf{Macro (\%)} & \textbf{Micro (\%)} & $\Delta$ \\
\midrule
Gemini-3.1-Pro & 47.8 & 47.1 & $-0.7$ \\
Gemini-3-Flash & 46.6 & 45.8 & $-0.8$ \\
Grok-4.2 & 44.9 & 44.2 & $-0.7$ \\
Kimi-K2.5 & 43.5 & 42.5 & $-1.0$ \\
Qwen3.5-397B & 43.4 & 42.5 & $-0.9$ \\
DeepSeek-V3 & 42.9 & 41.9 & $-1.0$ \\
Doubao-Seed-1.8 & 42.4 & 42.0 & $-0.4$ \\
GPT-5.2 & 35.8 & 35.0 & $-0.9$ \\
MiniMax-M2.5 & 35.2 & 34.4 & $-0.8$ \\
Claude-Opus-4.6 & 32.8 & 32.4 & $-0.4$ \\
MiMo-V2.5-Pro & 32.3 & 33.1 & $+0.8$ \\
Claude-Sonnet-4.6 & 30.8 & 30.3 & $-0.5$ \\
GPT-4o & 29.7 & 29.4 & $-0.3$ \\
\bottomrule
\end{tabular}
\end{table}

\section{Error Taxonomy}
\label{app:errors}
Table~\ref{tab:errors_full} gives the heuristic error-taxonomy distribution over 197{,}790 missed criteria across all 13 generators. Categories are assigned by parsing judge explanations for keyword patterns, then validated on a 50-sample manual audit (inter-rater agreement on category assignment was $>95\%$ on the audit subset; the keyword parsing on the full 197{,}790 was not human-validated end-to-end). The percentages should therefore be read as descriptive patterns in judge-explanation text, not as causal failure-mechanism counts.

\begin{table}[h]
\centering
\caption{Error taxonomy across all 13 generators (197{,}790 total misses).}
\label{tab:errors_full}
\small
\begin{tabular}{@{}lrr@{}}
\toprule
\rowcolor{CSBPanel}
\textbf{Error type} & \textbf{Count} & \textbf{\%} \\
\midrule
Missing specific reaction trigger (generic content) & 88{,}061 & 44.5 \\
Missing keyword vocabulary (T2) & 21{,}521 & 10.9 \\
Missing negative aspect (T4) & 20{,}438 & 10.3 \\
Meme repetition without pragmatic transfer & 17{,}375 & 8.8 \\
Missing positive aspect (T3) & 14{,}618 & 7.4 \\
Opposite valence (praise $\leftrightarrow$ criticism) & 14{,}112 & 7.1 \\
Missing flashpoint trigger name (T1) & 14{,}066 & 7.1 \\
Too generic / surface-level & 6{,}998 & 3.5 \\
Emotion match, wrong object & 372 & 0.2 \\
Factual error in comment & 229 & 0.1 \\
\bottomrule
\end{tabular}
\end{table}

\section{Why pointwise binary?}
\label{app:metric_ablation}

A natural question is whether the auditable pointwise judge is worth the engineering. We test the alternatives on a controlled 120-pair similarity benchmark (60 paraphrase positives, 30 random negatives, 30 hard negatives) constructed from released criterion examples. For each method, we score all pairs and compute AUC-ROC and best-threshold F1. The identical 0.667 Best-F1 values indicate a degenerate optimum: the best threshold predicts every pair as positive on the balanced 60/60 set.

\begin{table}[h]
\centering
\caption{Similarity-detection performance on 120 reaction-pair instances (60 positive, 60 negative). All four cheap unsupervised similarity baselines reach the trivial-baseline value in the \emph{Best F1} column (0.667 = $2/3$): at the optimal threshold they collapse to ``predict every pair as positive,'' which on a 60/60 split gives $P=0.5$, $R=1.0$, $F1=0.667$. LSA uses TF-IDF/SVD embeddings and cosine similarity. These baselines do not provide a usable replacement for pointwise judging.}
\label{tab:metric_ablation}
\small
\begin{tabular}{@{}lcc@{}}
\toprule
\rowcolor{CSBPanel}
\textbf{Method} & \textbf{AUC-ROC} & \textbf{Best F1} \\
\midrule
Token Jaccard (jieba-segmented) & 0.625 & 0.667 \\
Character N-gram Jaccard ($n=2$) & 0.618 & 0.667 \\
GPT-token Jaccard (\texttt{cl100k\_base}) & 0.604 & 0.667 \\
LSA embedding cosine (TF-IDF/SVD) & 0.608 & 0.667 \\
\bottomrule
\end{tabular}
\end{table}

Unsupervised similarity metrics fail for two reasons specific to consumer-reaction text: (i) paraphrase-positive pairs can share few surface units (e.g.\ ``feels like a corporate apology'' vs.\ ``reads like a PR statement''); (ii) hard-negative pairs can share many units but disagree on the criterion (e.g.\ ``reduced fireworks count'' as praise vs.\ ``fireworks were the problem to begin with'' as criticism). GPT-token overlap and LSA embeddings slightly raise AUC over raw surface overlap, but still collapse to the all-positive best-F1 threshold. This rules out cheap unsupervised similarity as a replacement for pointwise judging, while leaving learned cross-encoders as a possible future comparison.

But once a similarity \emph{must} be LLM-based, the next question is whether \emph{holistic} LLM-as-Judge (one prompt, one fuzzy quality score) is sufficient. On a 100+ criterion calibration set from 10 random topics, pointwise binary judging reaches \textbf{92.1\%} three-judge agreement, while holistic LLM-as-Judge on the same data reaches only \textbf{65.8\%}---a gap of nearly 30 points. Every additional point of judge disagreement contaminates downstream cross-model rankings, since the same generator can win or lose by judge variance alone. Pointwise binary judging is therefore the minimum sufficient design.

\paragraph{Why not a human-expert ceiling?} We do not treat human expert prediction as the target ceiling for \benchmark{}. The benchmark target is realized public discourse: comments that real consumers actually produced after seeing the event. Those comments can be surprising, messy, or difficult to infer, but their difficulty is exactly why a marketing-facing forecasting benchmark is needed. A PR or marketing expert baseline would be a useful diagnostic for expert foresight under the same brief, but it would not change the target definition. The goal is to recover concrete flashpoints and criticism vectors that later become salient in public discussion, including cases that human experts may also miss.

\section{Construction, Generation, and Judge Prompts}
\label{app:prompts}
\paragraph{Criteria drafting prompt.}
The operational drafting prompt was written in Chinese; below we give an abbreviated English translation. The drafting model receives topic metadata, an objective event description used as an anti-leakage reference, and filtered source reactions for the topic. It is instructed to produce benchmark ground truth as unambiguous scoring criteria rather than summaries.
\begin{quote}\small\ttfamily
You are a professional social-media public-opinion analyst. Your analysis will be used as ground truth for a public-reaction forecasting benchmark. The output must consist of unambiguous scoring criteria. Each criterion must have a clear boundary, positive examples, negative examples, and a one-sentence judgment rule, so that different raters can make consistent yes-no decisions.

Given the topic metadata, event description, and filtered source reactions, identify: (1) up to five emotion keywords with their emotion targets; (2) up to four positive aspects and up to four negative aspects; and (3) up to two sentiment flashpoints. Each item must be atomic and supported by the source reactions.

Anti-leakage rule: a criterion must capture user attitude, emotion, or viewpoint, not merely restate facts already present in the event description. If a generated comment only repeats names, times, places, actions, or outcomes from the event description without an attitudinal increment, it should not receive credit. Do not create a criterion if the source reactions do not provide enough support.

Return JSON with the four sections. For every criterion, include a name, definition, positive examples, negative examples, and judgment rule.
\end{quote}

\paragraph{Topic filtering prompt.}
Before criterion drafting, we use an LLM filter to retain hot topics related to brands, products, consumer behavior, or marketing-relevant service experiences. The operational prompt was written in Chinese; below is an abbreviated English translation.
\begin{quote}\small\ttfamily
Judge whether the following trending topic is related to brands, products, or commercial marketing.

Relevant types include: (1) brand events, such as releases, collaborations, controversies, or marketing campaigns; (2) product discussions, such as reviews, recommendations, warnings, or feature discussions; (3) consumer topics, such as consumption trends, shopping experience, and value-for-money discussion; (4) influencer commerce, such as creator recommendations, livestream commerce, and influencer-associated products; and (5) service experiences, such as restaurants, travel, medical, or other consumer-service experiences.

Exclude topics that are pure entertainment gossip without brand relevance, political news, sports events unless brand sponsorship is involved, and pure knowledge explainers.

Principle: prefer recall over precision; do not miss potentially relevant topics.

Trending topic: \{topic\}

Return JSON only: \{"related": true/false, "type": "...", "brand\_entity": "...", "reason": "short reason"\}.
\end{quote}

\paragraph{Semantic judging.} The judgment rules will not penalize models that capture a reaction's semantic essence with slight paraphrasing. This is because the judge prompt encodes five explicit semantic-equivalence rules, including emotion synonym/hierarchy matching, object-anchor matching, compound-emotion inclusion, strength versus factual statement, and meme object-transfer. A criterion such as ``\$2{,}000 starting-price leak'' is satisfied by any specific anchor in its judgment-rule range (``2{,}000 dollars,'' ``around 2k,'' ``\textyen 14{,}000+''), but not by generic phrasings such as ``it'll be expensive''. The latter does not reproduce the specific public-discourse anchor that drove the topic. The role of the lexical anchor is to distinguish a model that anticipates \emph{which framing the public actually chose} from one that emits any plausibly relevant comment; our held-out judgment-rule ablation and the per-criterion anatomy boxes in Appendix~\ref{app:casestudy} document this distinction.

\paragraph{Generator prompt.}
\begin{quote}\small\ttfamily
You are simulating realistic public reactions to a trending topic on Chinese social media. Topic: \{topic\_keyword\} | Event: \{event\_description\}. Generate \{k\} diverse, realistic comments that Chinese social-media users would post. Each should reflect a distinct perspective, emotion, or reaction. Write in natural Chinese social-media language.
\end{quote}
Comments per topic follow $k=\min(30,\max(10,|C_i|))$, where $|C_i|$ is the number of criteria for topic $i$.

\paragraph{Judge prompt.}
\begin{quote}\small\ttfamily
For each criterion, the judge receives the topic, the generated comments truncated to 3{,}600 characters, and the five-field criterion (name, definition, positive examples, negative examples, judgment rule) together with five explicit semantic-equivalence rules for Chinese: (1) emotion synonyms, (2) object matching, (3) compound inclusion, (4) strength vs.\ factual, (5) meme object-transfer. To keep each pointwise call bounded, the prompt includes up to five positive examples and three negative examples, each shortened to the first 120 characters; the release JSONL retains the full criterion metadata. The judge returns YES/NO with a short explanation. Full prompts are released with the benchmark.
\end{quote}

\section{Construction Hardening Details}
\label{app:hardening}
We manually check the drafted criteria and harden four recurring failure modes: event-restatement leakage, semantic duplication, over-pruning, and judge-fragile meme points. We use cosine similarity and LLM supervision to merge similar topics and criteria; drop criteria that collapse to the event description; verify within-topic independence with an LLM-assisted check; and audit pruned items so that meaningful but non-trivial criteria are restored rather than discarded. The hardening process ensures that simply repeating the task description earns no credit.

Meme-style criteria are a recurring failure mode for judges because a model can receive credit by simply repeating the meme template. We therefore rejudge 282 meme-like criteria across 225 topics, crediting only cases where the meme is transferred to a \emph{new object or scene} and forms a genuine second-order creation. For a real benchmark item, \emph{``School isn't basic, dorm isn't basic either''}, \textbf{(i)} \emph{``Face isn't basic, body isn't basic either, ability is even less basic''} transfers the template to a person and counts as a hit; \textbf{(ii)} \emph{``Dorm isn't basic, bathroom isn't basic either''} stays inside the original school/dorm domain and does not count; \textbf{(iii)} \emph{``Born not to be a basic model''} carries a similar meaning but does not use the structural template and also does not count.

To deal with over-pruning, we manually audit the pruned criteria. We restore 28 over-pruned criteria and revise another 87 into new ones. Table~\ref{tab:hardening_examples} shows real construction-log examples. We report only derived reaction signals, not raw user comments.

\begin{table}[h]
\centering
\caption{Examples of criterion hardening.}
\label{tab:hardening_examples}
\small
\renewcommand{\arraystretch}{1.12}
\begin{tabular}{@{}p{0.17\textwidth}p{0.20\textwidth}p{0.24\textwidth}p{0.29\textwidth}@{}}
\toprule
\rowcolor{CSBPanel}
\textbf{Topic} & \textbf{Failure mode} & \textbf{Draft signal} & \textbf{Hardened reaction signal} \\
\midrule
Aaaaapple apple juice police report & Event-restatement leakage & The brand formally filed a police report. & Replaced by stance-bearing reactions about legal-rights legitimacy, brand sincerity, celebrity effect, and heartache for the associated public figure. \\
Apple joins RedNote & Event-restatement leakage & Apple's official account joined RedNote. & Replaced by account-authenticity reversal and intellectual-property compliance, which require a user stance beyond the join fact. \\
Apple Watch product review & Semantic duplication and coarse labels & Repeated generic criteria such as anxiety, health-management value, aesthetic value, app ecosystem, durability, and comfort. & Split into higher-resolution signals about the ``beautiful but useless'' reversal, health-data visualization, abnormal battery drain, subscription-gated software, accessory quality, and resale-value reasoning. \\
iPhone 17 hands-on & Low-specificity launch hype & First-sale purchase enthusiasm. & Replaced by concrete discourse anchors such as eSIM-only policy, thinness versus battery capacity, durability concerns, purchase-channel advantage, and launch-policy confusion. \\
\bottomrule
\end{tabular}
\end{table}

\section{Extended Limitations}
\label{app:limitations}
This appendix expands on the caveats summarized in \S\ref{sec:release}.
\paragraph{(L1) LLM-judge bias and calibration scope.} The three-judge calibration (\S\ref{sec:judge}) and the three-annotator human audit provide evidence that the pointwise judge is reliable on a targeted calibration slice. This slice is designed as a judge-sensitivity stress test: criterion-level labels are compared across GPT, Gemini, Claude, and independent human-majority labels, rather than only by the leaderboard judge. The 3{,}600-character truncation reduces gross length bias, and criterion-specific definitions, examples, and negative examples reduce style-based drift. We therefore treat remaining judge ambiguity as a small semantic-boundary issue, not as a dominant source of the large model failures reported in the leaderboard.
\paragraph{(L2) Representative platform and portability.} The current release of \benchmark{} is built on Chinese RedNote (Xiaohongshu), a major consumer-facing UGC platform with 300M+ monthly active users and dense lifestyle, brand, product, entertainment, and public-event discourse. This makes RedNote a representative and practically important setting for consumer-behavior simulation. Many consumer reactions---price sensitivity, authenticity concerns, disappointment with over-marketing, praise for emotional value, criticism of service or design---are cross-platform phenomena. RedNote is also a demanding testbed: compared with many short-form platforms, its discourse often contains richer meme reuse, irony, self-presentation, and mixed affect, making it a strong stress test for consumer-reaction modeling. As a portability check, Appendix~\ref{app:youtube_pilot} instantiates the same construction on 100 English YouTube trending videos and observes a similar section-wise difficulty pattern. Future platform-specific releases can test how the same construction transfers to X, Reddit, B2B marketing, or other cultural settings, but \benchmark{} already covers a broad, commercially important consumer-reaction regime. Given the large gap to ceiling in Table~\ref{tab:leaderboard}, we view platform extension as an important future direction rather than the dominant explanation for current model failures.
\paragraph{(L3) Temporal contamination.} Source topics were surfaced between March and September 2025. Continuous pretraining or RLHF on newer frontier models may have ingested public traces of some events, so we do not claim contamination-free zero-shot forecasting for every proprietary model. However, seeing an event or scattered original comments is not enough to solve \benchmark{}. The model must compress its knowledge into a limited set of representative comments, while the score is computed against abstracted reaction criteria distilled from the full public-discourse evidence and hardened by our rules. These criteria are not raw comments copied into the prompt, and they remain hidden unless the benchmark itself leaks. Thus, contamination can still affect scores and rankings, but success requires recovering the aggregate reaction structure, such as stance, target, and judgment-rule match, rather than recalling isolated posts. We release timestamps and topic provenance to support future temporal splits and post-cutoff held-out evaluations.
\paragraph{(L4) LLM-assisted construction at benchmark scale.} Criterion drafting and initial judgments are LLM-assisted, followed by hardening, pruning, spot-checks, and human audit. This hybrid construction is the scalable design choice that makes a 23{,}122-criterion benchmark feasible while preserving inspectability. The human audit in \S\ref{sec:judge} compares pointwise judge decisions against independent human-majority labels on a held-out calibration slice, and the release includes every criterion's five-field metadata, provenance, and judgment rule so that users can inspect, filter, or extend the benchmark.
\paragraph{(L5) Endpoint coverage not all possible reactions.} \benchmark{} evaluates whether generated comments recover the concrete reaction points that actually surfaced in public discourse. This is the marketing-relevant target, where teams need to anticipate which criticisms, praise targets, emotional keywords, and flashpoints become salient enough to shape the public conversation before launching an official post or advertisement. The benchmark therefore does not credit every plausible but unobserved reaction equally, and it does not model the temporal process by which early reactions shape later ones. Dynamic opinion-diffusion simulation is complementary to the present benchmark.

\section{Second-Platform YouTube Pilot}
\label{app:youtube_pilot}
To test whether the construction procedure is portable beyond RedNote, we run a small second-platform pilot on English YouTube. We select 100 consumer-facing videos from the public Global YouTube Trending Dataset (2025 entries; English-speaking regions US/GB/CA/AU), balanced across 14 YouTube categories, and fetch the top 300 public comments per video through the official YouTube Data API. We then apply the same criterion-construction recipe: comments are abstracted into atomic reaction criteria with definitions, positive examples, negative examples, and judgment rules. The resulting pilot contains 1{,}508 criteria (mean 15.1 per topic) across 100 videos.

We evaluate the pilot with the same open-ended comment-generation interface and GPT-5.2 pointwise judge, using English prompts. The pilot is not part of the released leaderboard, but it provides a direct portability check. When moved from RedNote to English YouTube, the same endpoint-reaction formulation still exposes large gaps in recovering concrete public-discourse reactions. The section pattern also mirrors the main benchmark: positive aspects are easiest, while flashpoints and negative aspects remain much harder (Figure~\ref{fig:youtube_pilot}; Table~\ref{tab:youtube_pilot}).

\begin{figure}[h]
\centering
\includegraphics[width=0.95\textwidth]{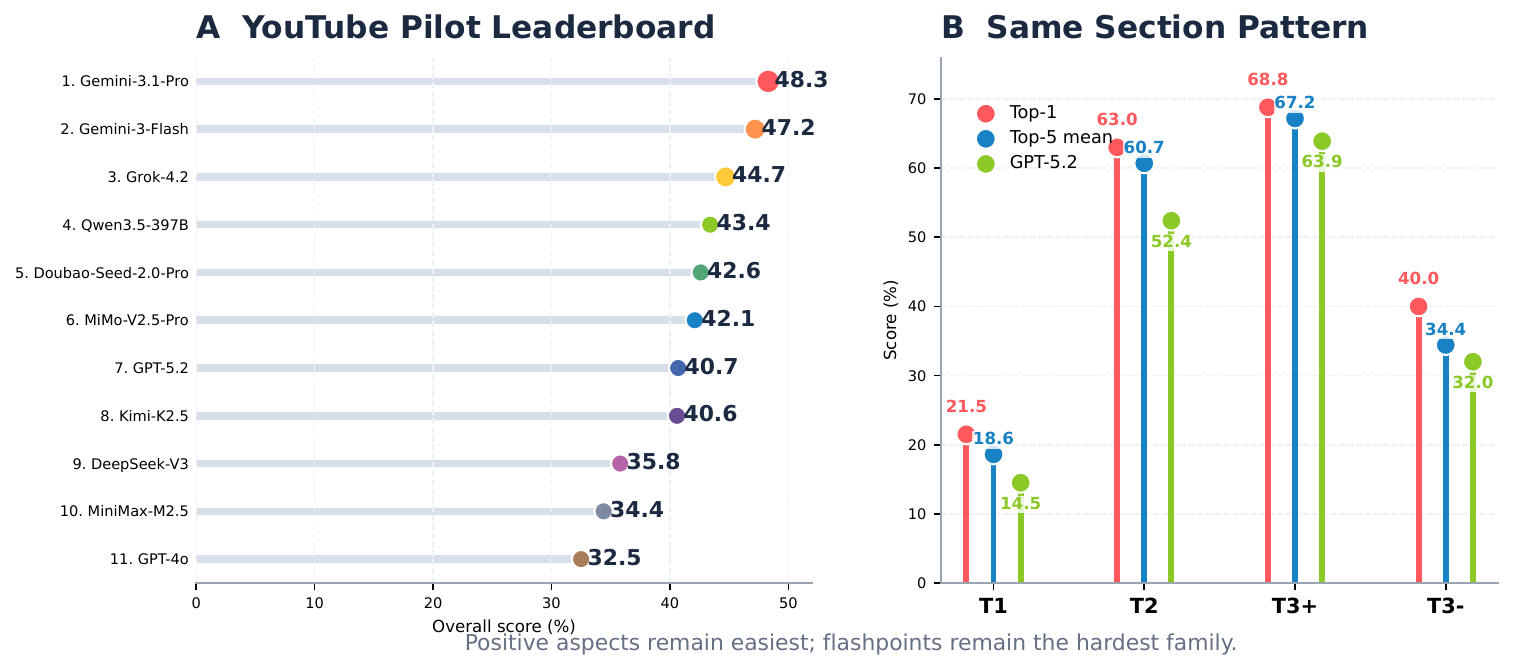}
\vspace{-0.8em}
\caption{Second-platform YouTube pilot. Left: overall lollipop leaderboard on 100 English trending videos. Right: section-wise pattern under the same endpoint-reaction protocol.}
\label{fig:youtube_pilot}
\vspace{-0.6em}
\end{figure}

\begin{table}[h]
\centering
\caption{Second-platform YouTube pilot on 100 English trending videos and 1{,}508 criteria. Scores are macro-averaged across topics under the same open-ended generation and pointwise judging protocol.}
\label{tab:youtube_pilot}
\small
\begin{tabular}{@{}lrrrrr@{}}
\toprule
\rowcolor{CSBPanel}
\textbf{Model} & \textbf{Overall} & \textbf{T1} & \textbf{T2} & \textbf{T3+} & \textbf{T3-} \\
\midrule
Gemini-3.1-Pro & 48.3 & 21.5 & 63.0 & 68.8 & 40.0 \\
Gemini-3-Flash & 47.2 & 21.2 & 60.3 & 71.0 & 36.2 \\
Grok-4.2 & 44.7 & 18.5 & 64.4 & 69.7 & 26.1 \\
Qwen3.5-397B & 43.4 & 15.9 & 57.2 & 63.7 & 36.7 \\
Doubao-Seed-2.0-Pro & 42.6 & 15.9 & 58.6 & 62.7 & 33.1 \\
MiMo-V2.5-Pro & 42.1 & 17.8 & 55.4 & 66.2 & 28.9 \\
GPT-5.2 & 40.7 & 14.5 & 52.4 & 63.9 & 32.0 \\
Kimi-K2.5 & 40.6 & 15.4 & 53.5 & 60.7 & 32.9 \\
DeepSeek-V3 & 35.8 & 12.5 & 45.8 & 55.5 & 29.5 \\
MiniMax-M2.5 & 34.4 & 11.7 & 45.2 & 57.4 & 23.4 \\
GPT-4o & 32.5 & 10.2 & 43.3 & 55.1 & 21.3 \\
\bottomrule
\end{tabular}
\end{table}

\section{Human Audit Protocol}
\label{app:human_audit}

The human audit in \S\ref{sec:judge} was designed to validate pointwise judge labels rather than to measure human forecasting ability. We sampled 100+ criterion-level items from 10 random topics, with the topic, event description, generated comments, and five-field criterion shown to each annotator. Three trained annotators independently assigned their own yes/no label indicating whether the generated comments satisfy the criterion. Each item also allowed an optional uncertainty flag for boundary cases. We then compared the frozen judge label with the human-majority label.

Annotators first completed a small calibration set with reference answers, then annotated fixed, independently shuffled versions of the formal audit set. The released annotation package includes the exact item files, annotation guidelines, exported anonymized labels, analysis script, and the single-file HTML annotation interface used in the study. The interface records only item-level decisions, timestamps, and optional uncertainty flags; it does not collect demographic, medical, behavioral, or other personal information from annotators. The audit was not run through a public crowdworking marketplace. Compensation and participation followed the applicable local/institutional protocol for this low-risk annotation task, and the released materials exclude payment records or identifying information to preserve anonymity.

\section{Reproducibility}
\label{app:repro}
All leaderboard evaluations use the same harness and fixed GPT-5.2 judge. Single generation call per topic (no best-of-$n$); generator temperature 0.7, max 4{,}096 tokens, the requested number of comments is $k = \min(30, \max(10, |C_i|))$. Judge temperature 0, max 4{,}096 tokens, one API call per atomic criterion. Total API cost is ${\sim}\$2{,}500$ across 13 generators and the judge. Wall-clock runtime for the main sweep is ${\sim}48$\,hours at 30 point workers $\times$ 8 topic workers. Scripts, config, and prompts are included in the release.

\section{Theoretical Foundations of the \model{} Framework}
\label{scf}
SCF (Strategy-Capital-Framing) is a classical consumer psychology framework that integrates three complementary social psychology theories to guide consumer behavior simulation.

\textbf{Strategy (S): Self-Presentation Theory.} Drawing from Goffman's dramaturgical analysis~\citep{goffman1959presentation}, we model consumers as strategic actors who manage impressions through their public statements. Key strategies include:
\begin{itemize}
    \item \textit{Ingratiation}: Expressing positive opinions to gain social approval
    \item \textit{Intimidation}: Asserting expertise or status through criticism
    \item \textit{Supplication}: Seeking sympathy through victimhood narratives
    \item \textit{Exemplification}: Demonstrating moral values through consumption choices
\end{itemize}

\textbf{Capital (C): Cultural Capital Theory.} Following Bourdieu~\citep{bourdieu1984distinction}, we analyze consumers' deployment of various capital types:
\begin{itemize}
    \item \textit{Economic capital}: References to price, value, affordability
    \item \textit{Cultural capital}: Displays of taste, aesthetics, expertise
    \item \textit{Social capital}: Name-dropping, community belonging
    \item \textit{Moral capital}: Ethical consumption, environmental consciousness
\end{itemize}

\textbf{Framing (F): Behavioral Economics.} Building on Prospect Theory~\citep{tversky1979prospect}, we attend to how consumers frame information:
\begin{itemize}
    \item \textit{Gain framing}: Emphasizing benefits and positive outcomes
    \item \textit{Loss framing}: Highlighting risks, regrets, or missed opportunities
    \item \textit{Reference points}: Comparisons to expectations, competitors, past experiences
\end{itemize}

\paragraph{Direct-generation prompt.}
The ablation in \S\ref{sec:scaffold-negative} used the following instruction before the standard generation request. The operational prompt was written in Chinese; we show an English translation here. The model was asked to perform the analysis internally and output only comments:
\begin{quote}\small\ttfamily
You are a real social-media user. Given the topic and event description, internally perform a Strategy--Capital--Framing analysis, but output only comments, not the analysis. For \textbf{Strategy}, consider why different users publicly participate: self-presentation through knowledge, taste, expertise, lived experience, or group identity; seeking resonance by linking the event to personal emotions, grievances, expectations, or anxiety; criticism or complaint about unreasonable, fake, awkward, or conflicted details; value signaling around morality, fairness, identity, stance, or boundaries; and meme-like joking through sarcasm, platform slang, remixing, or secondary creation. For \textbf{Capital}, cover possible identity and resource signals rather than generic emotion: cultural capital (aesthetic taste, knowledge, expertise, familiarity with the work, brand, or industry), social capital (fandom, city, profession, consumer group, or shared-experience belonging), economic capital (price sensitivity, value for money, consumption ability, budget pressure, expensive/cheap/worth-it judgments), and moral capital (empathy, justice, environmental concern, public good, respect for labor, anti-routine sentiment). For \textbf{Framing}, vary how users package the event: gain frames, loss frames, reference-point comparisons, identity frames, and spreadability frames that identify which details will be screenshotted, retold, memed, debated, or remixed. Generate the requested number of realistic comments, each 20--100 Chinese characters. Make comments concrete by using names, scenes, actions, prices, time, relationships, contrasts, or controversy points from the event; make comments clearly different rather than paraphrases; include support, skepticism, joking, empathy, extra information, personal experience, or risk warnings where natural; make them sound like casual user comments rather than analysis reports, and do not use the terms Strategy, Capital, or Framing.
\end{quote}

%%%%%%%%%%%%%%%%%%%%%%%%%%%%%%%%%%%%%%%%%%%%%%%%%%%%%%%%%%%%

\section{Case Study I: iPhone 17 Hands-On (Single-Model Per-Criterion Audit)}
\label{app:casestudy}

This case study examines a single benchmark instance under the leaderboard's \#1 model (Gemini-3.1-Pro), exhaustively annotating every ground-truth criterion with a \cmark/\xmark and a pointer to the specific generated comment that satisfies it. The topic is \emph{iPhone~17 Hands-On}: a high-engagement consumer-tech thread with 68 atomic criteria (T1=14, T2=24, T3=12, T4=18) drawn from real RedNote reactions, and a generation budget of 20 comments. With full event briefing in the prompt, a model must anticipate \emph{which specific reaction vectors will dominate the public discussion}---the difference between ``Apple released a phone'' and ``the rumored \$2{,}000 starting price, two-tone seam aesthetic, and foldable-roadmap lag are the three vectors that will set the public mood''---rather than merely produce plausible commentary about a phone launch.

\paragraph{Headline scores on this topic.}
Final 25.7\%, decomposed as T1 = 1/14 (7\%), T2 = 9/24 (37\%), T3 = 3/12 (25\%), T4 = 6/18 (33\%). The model is fluent and emotionally varied, but its T1 coverage collapses, missing 13 of 14 flashpoints. This single instance reproduces, locally, the asymmetric-failure pattern observed across the full benchmark.

\paragraph{Topic and event description (the prompt the model receives).}
\emph{Apple is preparing a launch event for the iPhone~17 series. Industry analysts and online leaks indicate hardware upgrades across the lineup: a 24~MP front camera on all models; a 48~MP telephoto lens on the Pro variants; 12~GB unified memory; the new A19 and A19 Pro chips on TSMC's 3~nm process. High-refresh-rate displays are expected across the line, with possible improvements to battery life and charging speed. Event date rumored for September.}

\paragraph{Sample generated comments (translated verbatim; 8 of 20).}
\begin{quote}\small
\textbf{C1.} 12~GB unified RAM across the line? Apple's finally being generous! Used to be so stingy. Now I won't have to worry about background apps getting killed. Speaking as an iPhone~13 holdout---this time I really am ready to upgrade.\\[0.3em]
\textbf{C2.} The base model finally gets a high-refresh display? In what era? Even budget Android phones have had this for years; Apple really squeezes feature upgrades drop by drop. But better late than never---hopefully no more 60~Hz eye-strain.\\[0.3em]
\textbf{C4.} A19 and A19~Pro on TSMC's 3~nm process---performance will obviously be top-tier. But honestly, the A-series chips have been over-spec'd for a while. What I actually care about is whether the thermals can keep up; don't want gameplay throttling and screen-dimming again.\\[0.3em]
\textbf{C7.} Battery and fast-charge upgrades are rumored, but Apple's ``fast charging'' is what it is---reaching 40~W would be a miracle. Domestic Android phones start at 120~W. Apple really needs to step up here.\\[0.3em]
\textbf{C9.} Looking at the renderings and so-called hands-on photos online, the appearance basically hasn't changed---same body, new color, sold for another year. Phones have hit a plateau; even Apple can't produce anything visually new anymore.\\[0.3em]
\textbf{C12.} RAM is going up, but what about storage? Don't tell me the base model still starts at 128~GB. A few WeChat groups and games and you're full. 256~GB minimum is the only configuration that makes sense in 2025.\\[0.3em]
\textbf{C16.} Complain all you want, but you'll buy it anyway. Full-stack Apple users have nowhere to go---Mac, iPad, Watch all locked together. As long as iPhone 17 doesn't crash and burn, of course we'll upgrade. This is what ecosystem lock-in looks like.\\[0.3em]
\textbf{C19.} Whatever the chip and camera, can they please fix the signal first? Step into an elevator or a parking garage and lose service. Try scanning a shared bike and it won't read. \emph{This} is the daily-use pain point that actually matters.
\end{quote}
The output reads natural, varied in stance (early-adopter, fence-sitter, holdout, ecosystem-trapped pragmatist), and surfaces several legitimate reactions: innovation fatigue, surprise at the high-refresh upgrade, charging-speed grievance, signal-quality complaint. The audit reveals where it falls short.

\paragraph{What the audit catches.}
Table~\ref{tab:case_iphone} shows a representative slice of the per-criterion judge decisions, with a ``Evidence'' column pointing to the specific generated comment that satisfies the criterion (or ``---'' for misses). The pattern is clear. The model scores well on \emph{generic} reactions (excitement, ``feels incremental,'' praise for the camera bump) and fails on \emph{specific cultural anchors} that drove the actual public discourse.

\begin{table}[!htbp]
\centering
\caption{Per-criterion judge decisions on \emph{iPhone~17 Hands-On}, illustrative subset. \cmark = covered, \xmark = missed. ``Evidence'' points to the specific generated comment that satisfies the criterion (or ``---'' for a miss).}
\label{tab:case_iphone}
\small
\renewcommand{\arraystretch}{1.10}
\begin{tabular}{@{}p{0.05\columnwidth}p{0.25\columnwidth}p{0.04\columnwidth}p{0.06\columnwidth}p{0.50\columnwidth}@{}}
\toprule
\rowcolor{CSBPanel}
\textbf{Sec} & \textbf{Criterion} & & \textbf{Ev.} & \textbf{Judge reasoning (condensed)} \\
\midrule
\rowcolor{CSBPanel}
\multicolumn{5}{@{}l}{\textbf{\textcolor{CSBInk}{T1. Sentiment flashpoints}}} \\
T1 & ``\$2{,}000 starting-price leak'' & \xmark & --- & Comments voice ``probably will go up,'' but never quote the specific price anchor that the actual discourse centered on. \\
T1 & ``5.5~mm thinness vs.\ 2900~mAh battery contradiction'' & \xmark & --- & Battery (C7, C17) and thinness are each mentioned but never juxtaposed as a trade-off. \\
T1 & ``Pro two-tone seam aesthetic backlash'' & \xmark & --- & C9 says ``looks the same''; the bimaterial / glass-aluminum seam is never named. \\
T1 & ``A19 chip on TSMC 3~nm'' & \cmark & C4 & Comment 4 explicitly names ``A19 / A19~Pro on TSMC 3~nm.'' \\
\addlinespace[0.2em]
\rowcolor{CSBPanel}
\multicolumn{5}{@{}l}{\textbf{\textcolor{CSBInk}{T2. Emotion keywords}}} \\
T2 & ``Excitement'' & \cmark & C1, C5 & Strong positive intensifiers (``finally,'' ``alarm set,'' ``ready to upgrade''). \\
T2 & ``Pleasant surprise at features'' & \cmark & C1, C2 & 12~GB RAM and base-model high-refresh are framed with surprise markers. \\
T2 & ``Wavering / brand-switch consideration'' & \xmark & --- & C16 explicitly stays inside the Apple ecosystem (``ecosystem lock-in''). \\
T2 & ``Confusion over eSIM-only China delay'' & \xmark & --- & The eSIM regulatory issue is never raised. \\
\addlinespace[0.2em]
\rowcolor{CSBPanel}
\multicolumn{5}{@{}l}{\textbf{\textcolor{CSBInk}{T3. Positive aspects}}} \\
T3 & ``Imaging system upgrade'' & \cmark & C3, C6 & 24~MP front and 48~MP Pro telephoto are praised. \\
T3 & ``eSIM technology push'' & \xmark & --- & No comment positively frames eSIM as space-saving or industry-progressive. \\
T3 & ``Storage policy sincerity (256~GB base)'' & \xmark & --- & C12 demands 256~GB but never credits Apple for an actual policy change. \\
\addlinespace[0.2em]
\rowcolor{CSBPanel}
\multicolumn{5}{@{}l}{\textbf{\textcolor{CSBInk}{T4. Negative aspects}}} \\
T4 & ``Innovation fatigue'' & \cmark & C9 & C9: ``hit a plateau, even Apple can't produce anything visually new.'' \\
T4 & ``Incrementalism (`squeezing toothpaste')'' & \cmark & C2 & C2: ``squeezes feature upgrades drop by drop.'' \\
T4 & ``China-region feature limits (AI / eSIM)'' & \xmark & --- & C19 names a generic ``signal'' complaint, not the China-SKU restriction. \\
T4 & ``Foldable-roadmap lag vs.\ Android'' & \xmark & --- & No mention of Apple being late to the foldable market. \\
\bottomrule
\end{tabular}
\end{table}

\paragraph{Anatomy of one decision: the full audit trail.}
To make the audit fully inspectable, we expose the complete five-field structure used by the judge for two representative criteria---one missed, one covered.

\casebox{
\textbf{T1 ``\$2{,}000 starting-price leak'' \xmark\ (missed)}\\
\emph{Definition.} Discussion of the rumored \$2{,}000 / RMB 14k+ starting price.\\
\emph{Positive examples.} ``Starting price will reach \$2{,}299 (about RMB 16{,}635).''\quad ``Pricing might be in the \$1{,}800--\$2{,}000 range.''\quad ``Could go as high as \$2{,}500.''\\
\emph{Negative examples.} ``Bill of materials is \$759.'' (cost, not price)\quad ``Used Find~N3 around 5{,}000.'' (competitor price)\\
\emph{Judgment rule.} Must directly mention a specific price prediction near \$2{,}000 \emph{or} above RMB 14{,}000.\\
\emph{Judge decision and reasoning.} \xmark\ Comments mention ``don't be too pricey'' / ``Cook please go easy on us'' (C8), but never quote the specific \$2{,}000 / 14k anchor; criterion not met.
}

\casebox{
\textbf{T1 ``A19 chip on TSMC 3~nm process'' \cmark\ (covered)}\\
\emph{Definition.} Discussion of the new A19 / A19~Pro chip and its 3~nm fabrication process.\\
\emph{Positive examples.} ``A19 / A19~Pro on TSMC 3~nm.''\quad ``A19 hits 3~nm at last.''\\
\emph{Negative examples.} ``A19~Pro will use TSMC 2~nm'' (wrong process node).\\
\emph{Judgment rule.} Must mention ``A19'' \emph{and} the 3~nm process, with stance toward performance / thermals.\\
\emph{Judge decision and reasoning.} \cmark\ Comment~4 reads ``A19 and A19~Pro on TSMC's 3~nm process---performance will obviously be top-tier. But honestly, the A-series chips have been over-spec'd for a while; what I care about is thermals.'' Names both anchors and adds a thermal-stance, satisfies the rule.
}

% \paragraph{What two more deep-dives reveal: hits driven by lexical anchor, misses driven by missing cultural anchor.}
% The full audit trail above generalizes. Across the 14 T1 flashpoint criteria, every \cmark\ comes from a comment that reproduces the specific lexical anchor named in the criterion's judgment rule (``A19,'' ``3~nm,'' ``squeezing toothpaste,'' ``alarm set''). Every \xmark\ is a case where the model produces fluent commentary on the same broad topic but does not retrieve or invent the specific anchor (``\$2{,}000 starting price,'' ``two-tone seam,'' ``2900~mAh battery vs.\ 5.5~mm''). The audit thus localizes the failure mechanism: it is not that the model lacks emotional range, fluency, or topical relevance---it is that it cannot anticipate \emph{which specific framing the public has chosen}, and so cannot generate it on demand from an event description alone.

\paragraph{Why the model fails: a forecasting problem, not a fluency problem.}
The misses cluster on items that require knowing \emph{which specific narrative} is attaching itself to this product cycle, not just what the product is. The \$2{,}000 price anchor, the seam-aesthetic backlash, the foldable-roadmap delay, and the China-region eSIM rollout caveat are all real flashpoints in the actual RedNote discussion---each documented across hundreds of posts. The event description names the launch and the headline specs, and the model is expected to combine that brief with its prior knowledge of how RedNote consumers historically frame Apple launches to anticipate the specific framings above. The model's actual generation handles the easier task (``predict that people will discuss innovation, refresh rate, charging speed'') competently, while failing the harder task (``predict that the dominant negative thread will be the seam aesthetic, not the chip generation'').

\paragraph{Takeouts.}
\textbf{(i)~Difficulty is local, not just statistical.} Even on one high-attention topic, the asymmetric pattern (T1 collapses, T2 partial, T3/T4 partial) reproduces the global Table~\ref{tab:leaderboard} pattern.
\textbf{(ii)~Audit decomposes a single fuzzy quality into 68 inspectable decisions.} A reviewer or practitioner can verify each \cmark / \xmark independently against the criterion's definition, examples, and judgment rule, instead of accepting an opaque ``25.7\% similarity.''
\textbf{(iii)~The hard sections are the operationally relevant ones.} For pre-launch risk assessment and crisis forecasting, T1 (which trigger goes viral) and T4 (which complaint dominates) are the load-bearing signals; high T2 emotion-vocabulary coverage with low T1 flashpoint coverage---as observed here---would have limited use for these applications, even at high headline scores.

\FloatBarrier

\section{Case Study II: Aaaaapple Apple Juice (Cross-Model Coverage Comparison)}
\label{app:case_apple_juice}

This case study examines a single topic across five generators in order to characterize how the leaderboard's per-model gap appears at the level of individual generated sentences. The topic, \emph{Aaaaapple Apple Juice files police report}, is a topic on which the per-model spread reaches 30+ points (final scores 0.30--0.60 across our sample of five models), one of the largest spreads observed among brand-relevant topics, while the typical model spread is closer to 8--12 points (Table~\ref{tab:leaderboard} confidence intervals). The topic was chosen for diagnostic resolution. All five models receive the same event description and none has an obvious knowledge advantage---the brand and celebrity are real but neither is internationally famous---so the score gradient isolates differences in \emph{stance distribution}, \emph{first-person voice}, and \emph{register occupied}. On this topic, final scores are: Grok-4.2 0.60, Gemini-3.1-Pro 0.55, Claude-Sonnet-4.6 0.32, GPT-4o 0.30, GPT-5.2 0.29.

\paragraph{Topic and event description.}
\emph{``Aaaaapple'' is a brand selling apple juice and apple chips. The brand recently became viral after being associated with celebrity Zhao Lusi, who reportedly endorsed it during a rural-farming charitable initiative. After accusations spread online (some users alleged the celebrity-tie was fake or exploitative), the brand publicly announced it had filed a police report and preserved evidence against 11 specific bloggers. The incident sits at the intersection of celebrity endorsement, rural-aid marketing, and a contested PR response.}

\paragraph{Challenges in this topic.}
The 26 ground-truth criteria are unusually narrative-specific: covering them requires (i) knowing the celebrity-endorsement backstory, (ii) recognizing the ``hailstorm-damaged apple farming-help'' framing that drove the story to virality, and (iii) tracking emotional ambivalence (sympathy for the brand vs.\ skepticism about commercialization of charity). A generic ``brand files police report'' generation gets nowhere, while a generation that engages the celebrity-charity-skepticism triangle scores much higher.

\paragraph{Side-by-side model output (opening 2--3 comments per model).}
We sample three models---a strong, mid-tier, and weak generator---to make the differentiation visible. All three were given the same event description.

\casebox{
\textbf{Gemini-3.1-Pro (strongest, final 0.55):}\\
\emph{C2 (Defending celebrity, fan-adjacent stance).} Random bystander here---Zhao Lusi was just doing a small farming-help charity. Why is everyone dumping dirt on her? The brand was right to file the report; hopefully the police can clear her name. Don't let people who do charity get burned for it.\\[0.3em]
\emph{C5 (Product-quality defense vector).} Setting aside the gossip---their apple chips are actually pretty good, the ingredient list is clean. Hopefully the brand can put more energy into product quality and not let online drama ruin a good reputation.
}

\casebox{
\textbf{Claude-Sonnet-4.6 (mid-tier, final 0.32):}\\
\emph{C3.} The brand reporting 11 bloggers is genuinely shocking. Online speech freedom is precious, but it's not a license for defamation or rumor-spreading. Hopefully this case becomes a warning---spread information freely, but also responsibly.\\[0.3em]
\emph{C4.} Seeing the brand take legal action to defend its rights, I think this is the right choice for a modern enterprise facing the online-opinion environment. Cyberspace is not a lawless zone.
}

\casebox{
\textbf{GPT-4o (weak-tier, final 0.30):}\\
\emph{C2.} I think Aaaaapple's reaction is a bit excessive---isn't user discussion also free promotion for the brand? Though defending the brand image is also understandable.\\[0.3em]
\emph{C4.} Support Aaaaapple defending its rights! The internet is not a lawless zone; rumor-mongers should bear responsibility.
}

\paragraph{What changes across models.}
Three observations track the score gap. \textbf{(i)~Stance diversity.} Gemini's five comments occupy distinct positions (defender, fan-adjacent, onlooker, marketing-stunt skeptic, product-focused); Claude's are all measured and procedural (``cyberspace is not a lawless zone'' appears in 3 of 5); GPT-4o's are short neutral takes that all lean ``support the brand.'' \textbf{(ii)~Affective register.} Gemini surfaces fan-protective vocabulary (``don't let people who do charity get burned''); Claude and GPT-4o use abstract legal-procedural language. \textbf{(iii)~Domain pivot.} Gemini explicitly pivots to product quality (apple chips, clean ingredient list); the others stay event-meta. The score gap is built almost entirely from these three differences in \emph{how} the reactions are framed, not what facts the model knows.

\paragraph{Per-model diagnosis: why each model lands where it does.}
Reading the actual generations alongside the per-criterion judge decisions, each model's score is explainable. We give each generator a one-paragraph diagnosis with quoted evidence, stance counts, and section-level breakdown.

\noindent\textbf{Grok-4.2 (final 0.60; T1 = 2/3, T2 = 7/10, T3 = 5/6, T4 = 2/7).}
\emph{Win condition: stance breadth.} Across its 15 comments we count five distinct stances: fan-defender (C1, C2, C5), skeptic-comedy (C3 ``you're selling apple juice or scripts?,'' watermelon-eating emoji), concerned-consumer (C4), hater-blaming (C5: ``Zhao Lusi's haters are at it again''), and brand-policy commentator. Comedic/meme register (emoji-heavy, mock-product-as-knockoff-Apple-phone joke) is a stance no other model occupies. This breadth is exactly the registers that T2 emotion-keyword criteria require. Loses on T4 (2/7) because none of the comedic or fan-defender stances surface specific product-criticism vectors (portion size, packaging defects, customer service)---i.e., the same universal-miss class as the other 4 models.

\noindent\textbf{Gemini-3.1-Pro (final 0.55; T1 = 2/3, T2 = 6/10, T3 = 4/6, T4 = 2/7).}
\emph{Win condition: first-person stance occupation.} 4 of the 5 opening comments name an explicit speaker position before stating an opinion: ``a passing bystander says...'' (C2), ``setting aside the gossip...'' (C5), ``only I think this looks like a coordinated marketing play?'' (C4), ``the melon-eaters are confused'' (C3). This explicit positioning is what the \emph{xin-teng} (heartache) and \emph{lien-ai} (tender concern) criteria require: those criteria are scored only when the comment expresses sympathy from a recognisable vantage point (a fan, a sympathetic bystander), not as floating commentary. Gemini also pivots to product quality in C5 (``apple chips are good, ingredient list is clean''), uniquely among the five models, satisfying the T3 product-flavor criterion. Loses on T4 for the same reason as Grok: no model surfaces concrete consumer complaints.

\noindent\textbf{Claude-Sonnet-4.6 (final 0.32; T1 = 2/3, T2 = 4/10, T3 = 3/6, T4 = 0/7).}
\emph{Loss mechanism: stance-mode collapse onto institutional-legal.} 4 of the first 5 comments share the same structural template: ``[event statement] $\to$ rights-justifying claim ($\to$ aspiration about responsible online discourse).'' The phrase ``cyberspace is not a lawless zone'' or its near-paraphrase appears in C2, C3, C4 (and a near-variant in C6). The model produces fluent text, but the stance distribution is narrow: the comment fan-out concentrates on one register---the institutional-legal defender. Specific affective vantages (sympathetic fan, marketing-stunt skeptic, product-defender, hater-blaming) are absent, so 6 T2 affective criteria miss. T4 = 0/7 is the most telling number: the legal-defense stance generates no critical content, even though T4 is exactly the negative-aspect content a marketing application targets. This pattern is consistent with Claude-Sonnet's overall 30.8\% (rank 12/13).

\noindent\textbf{GPT-4o (final 0.30; T1 = 2/3, T2 = 4/10, T3 = 2/6, T4 = 1/7).}
\emph{Loss mechanism: brevity + neutrality.} Comment length: 21--34 Chinese characters (vs 80--140 for Gemini, 60--100 for Grok). Short comments cannot anchor specific stances. Balance markers (``a bit excessive, but understandable'' C2; ``hope both sides resolve peacefully'' C6) further wash out stance signal. The result is comments that read like neutral mediator statements, satisfying neither the heartache criterion (which requires a sympathetic stance toward the target) nor the marketing-stunt criterion (which requires a skeptical stance). This mirrors the global pattern: GPT-4o's overall 29.7\% (rank 13/13) is built from short balanced output, not from knowledge gaps---it knows the event facts but cannot produce the stance-occupying generations the benchmark requires.

\noindent\textbf{GPT-5.2 (final 0.29, lowest; T1 = 1/3, T2 = 3/10, T3 = 4/6, T4 = 1/7).}
\emph{Loss mechanism: meta-analytical bias (journalist register).} Comments read like opinion-column analysis: ``if someone really impersonated the celebrity for fake endorsements, accountability is necessary'' (C3); ``maybe the brand should clarify which information is false'' (C5); ``11 bloggers being legally pinned suggests this isn't ordinary criticism'' (C6). Every comment occupies the same meta-strategic register---a journalist or analyst, not a consumer. The first-person consumer voice (``I bought to help,'' ``it tastes good,'' ``Lusi got smeared, my heart goes out'') is structurally absent. Result: T2 affective criteria (\emph{lien-ai}, \emph{xin-teng}) systematically miss; T3 product-experience criteria miss. Notably T1 = 1/3 is also the lowest, showing that even on factual-anchor criteria the journalist register hurts---summary commentary doesn't reproduce specific anchors. This pattern recurs globally: GPT-5.2's 35.8\% versus Gemini's 47.8\% gap is dominated by T2 (35.1\% vs 56.9\%, 22 points) and T3 (47.8\% vs 61.2\%, 13 points)---exactly the registers a meta-analytical voice cannot produce.

\paragraph{Criterion construction and validity.}
The hard-tail criteria are deliberately specific because they represent the concrete anchors that made the real discourse recognizable. \benchmark{} evaluates whether a model can forecast real public-discourse reactions, not merely produce reactions entailed by a short event description. The comments that define these criteria were themselves produced by consumers reacting to the event, so removing them would erase exactly the signals that a marketing-facing simulation system should recover. A five-field record lets the judge distinguish generic near-misses from genuine coverage. Thus, a miss indicates failure to reconstruct a reaction pattern present in the source discourse.

\begin{table}[!htbp]
\centering
\caption{Per-criterion coverage on \emph{Apple Juice files police report}, illustrative subset across 5 representative models. \cmark\ = covered, \xmark\ = missed. The pattern: stronger models pick up criteria requiring the celebrity-charity context; weaker models stay at generic brand-crisis language.}
\label{tab:case_apple_juice}
\small
\renewcommand{\arraystretch}{1.10}
\setlength{\tabcolsep}{4pt}
\begin{tabular}{@{}>{\centering\arraybackslash}p{0.025\textwidth}p{0.21\textwidth}*{5}{>{\centering\arraybackslash}p{0.04\textwidth}}p{0.42\textwidth}@{}}
\toprule
\rowcolor{CSBPanel}
\textbf{S} & \textbf{Criterion} & \textbf{Gem.} & \textbf{Grok} & \textbf{GPT-5.2} & \textbf{GPT-4o} & \textbf{Son.} & \textbf{Why models split} \\
\midrule
\rowcolor{CSBPanel}
\multicolumn{8}{@{}l}{\textbf{\textcolor{CSBInk}{T1. Sentiment flashpoints}}} \\
T1 & ``Charity-intent maliciously questioned'' & \cmark & \cmark & \xmark & \xmark & \xmark & Requires linking the celebrity's farming-aid intent to public skepticism; weaker models stop at ``defamation lawsuit.'' \\
T1 & ``Hailstorm-damaged apple farming-help backstory'' & \xmark & \xmark & \xmark & \xmark & \xmark & Universal miss: the specific ``hail-damaged crop'' framing is the actual viral hook; no model surfaces it from the event brief plus its prior knowledge of how RedNote consumers frame brand--charity stories. \\
\addlinespace[0.2em]
\rowcolor{CSBPanel}
\multicolumn{8}{@{}l}{\textbf{\textcolor{CSBInk}{T2. Emotion keywords}}} \\
T2 & ``Sympathy / heartache for brand'' & \cmark & \xmark & \xmark & \xmark & \xmark & Only Gemini surfaces the protective-sympathy emotion; others default to neutral commentary. \\
T2 & ``Tender concern (\emph{lian-ai})'' & \cmark & \cmark & \xmark & \xmark & \xmark & Stronger models capture the celebrity-fan vocabulary; weaker models use generic emotion words. \\
T2 & ``Altruistic pride (`I helped a farmer')'' & \xmark & \xmark & \xmark & \xmark & \xmark & Universal miss: the first-person ``I bought to help'' frame is dominant in the actual discourse, but absent in all generations. \\
\addlinespace[0.2em]
\rowcolor{CSBPanel}
\multicolumn{8}{@{}l}{\textbf{\textcolor{CSBInk}{T3. Positive aspects}}} \\
T3 & ``Product flavor'' & \cmark & \cmark & \xmark & \xmark & \xmark & Stronger models include taste praise, anchoring brand-defense in product quality; weaker models stay event-meta. \\
T3 & ``Brand thoughtfulness (gifting / packaging)'' & \xmark & \xmark & \xmark & \xmark & \xmark & Universal miss: no model anticipates that customer-service / packaging praise becomes a defense vector when a brand is under attack. \\
\addlinespace[0.2em]
\rowcolor{CSBPanel}
\multicolumn{8}{@{}l}{\textbf{\textcolor{CSBInk}{T4. Negative aspects}}} \\
T4 & ``Online discourse environment deterioration'' & \cmark & \xmark & \xmark & \xmark & \xmark & Only Gemini connects the police-report move to a meta-narrative about toxic online behavior. \\
T4 & ``Product portion / quality-control complaints'' & \xmark & \xmark & \xmark & \xmark & \xmark & Universal miss: actual users post specific complaints (``shards too small,'' ``portion light''); no model generates these unprompted. \\
T4 & ``Customer-service experience'' & \xmark & \xmark & \xmark & \xmark & \xmark & Universal miss: scripted-response complaints are operationally important but never appear in generated comments. \\
\bottomrule
\end{tabular}
\vspace{-2em}
\end{table}

\paragraph{Why stronger models pick up what weaker models miss.}
The covered-by-Gemini-only criteria share a common pattern: they are \emph{specific affective stances}, not events. ``Heartache for the brand'' (\emph{xin-teng}), ``tender concern'' (\emph{lian-ai}), ``online environment deterioration'' --- each requires the model to occupy a particular emotional vantage point (a fan defending the celebrity, a sympathetic third party, a meta-observer of the discourse), then write from that vantage. Weaker models default to neutral, structurally similar comments (``I'll wait for police results,'' ``hope the brand is treated fairly''). This is consistent with the global finding that Gemini-3.1-Pro's edge over GPT-5.2 / Claude is concentrated in T2 emotion-vocabulary breadth and T3 affective specificity. The evidence is more consistent with affective- and stance-range limitations than with simple event-fact ignorance.

\paragraph{Anatomy of a model-discriminating criterion.}
The same 5-field structure as Case~I applies; we expose one criterion to make the discrimination concrete.

\casebox{
\textbf{T2 ``Heartache (\emph{xin-teng}) for celebrity / brand''}\\
\emph{Definition.} Sympathy and felt-injustice on behalf of the celebrity or brand under online attack.\\
\emph{Emotion target.} Zhao Lusi (the celebrity) / brand owner.\\
\emph{Positive examples.} ``Lusi probably thought this was just a quick favor, but ended up spending huge time fixing things.''\quad ``Being smeared, doubted, framed, maliciously redirected---stay away from that toilet of a platform.''\quad ``Lusi quietly built this since the start of the year; only now we realize.''\\
\emph{Negative examples.} ``I want to praise Lusi to death!'' (this is praise, not heartache)\quad ``Good people get good karma'' (blessing, no acknowledgement of harm).\\
\emph{Judgment rule.} Must explicitly mention negative treatment received (criticism, rumors, hardship) and pair it with a sympathetic affective response.\\[0.4em]
\textbf{Decisions across the 5 sampled models:}\\
$\bullet$ \textbf{Gemini-3.1-Pro} \cmark: C2 ``random bystander here---don't let people who do charity get burned for it'' explicitly names the negative treatment + sympathetic stance.\\
$\bullet$ \textbf{Grok-4.2} \xmark: comments express skepticism or fan-loyalty but never the heartache stance with negative-treatment framing.\\
$\bullet$ \textbf{GPT-5.2} \xmark: comments stay procedural / strategic; no felt-injustice register.\\
$\bullet$ \textbf{Claude-Sonnet-4.6} \xmark: ``hopefully this becomes a warning'' is meta-procedural, not sympathetic to the target.\\
$\bullet$ \textbf{GPT-4o} \xmark: short comments lean either ``brand should defend itself'' or ``brand is over-reacting''; neither occupies the heartache vantage.
}

\paragraph{Takeaways.}
\textbf{(i)~The strong-model edge is affective range, not knowledge.} All five models received the same event description and produced topical, fluent generations. The score gap traces to whether the model can occupy specific emotional vantage points (fan-defender, sympathetic third party, marketing-stunt skeptic) or defaults to procedural / abstract commentary.
\textbf{(ii)~Auditable scoring exposes which stance is missing.} Without per-criterion judgment, we would only know that ``Gemini scored higher.'' With the audit, we can name the specific affective register (heartache, tender concern, online-environment criticism) that the weaker models missed.
\textbf{(iii)~Operationally relevant gap.} A marketing system that produces only procedural / legal-defense comments cannot model what fans, skeptics, and meta-observers will actually say---which is the core forecasting requirement.

\paragraph{What both strong and weak models miss.}
The hardest tail criteria in this topic --- the near-universal hailstorm-damaged apple framing, plus zero-coverage altruistic-pride first-person voice, brand-thoughtfulness defense vector, and product-QC complaints --- form a coherent class. They are all \emph{anchored in concrete consumer experience} (a real story, a real personal narrative, a real product complaint), not in abstract event reasoning. Anticipating them given the event description requires bringing prior knowledge of how RedNote consumers actually talk about brand crises---knowledge that the benchmark targets but almost no current generator surfaces.

\FloatBarrier

\FloatBarrier

\section{Case Study III: Difficult Tail Criteria (Cross-Topic Frontier Analysis)}
\label{app:universal_misses}

This case study characterizes the hardest tail of the benchmark: criteria recovered by no model, plus a small set recovered by only one model. Whereas Cases~I and II diagnose model-specific failure modes on individual topics, Case~III first constructs the universal-miss set by intersecting per-model coverage decisions across all 13 generators and isolating criteria with zero hits. On the final benchmark, \textbf{6{,}677 of 23{,}122 atomic criteria (28.9\%)} receive zero coverage. We separately inspect 152 near-universal misses covered only by MiMo-V2.5-Pro, because they reveal what a single model can sometimes recover while the rest of the frontier misses. To control for topic-specific or section-specific artifacts, the gallery below samples across topics and across all four reaction families; included criteria carry the full 5-field structure with abstracted positive examples rather than released raw user threads.

The missing criteria cluster around discourse-latent framings: cultural anchors (e.g., the 24-solar-term seasonal calendar), viral meme templates, first-person consumer voice (\emph{``I bought to help''}), and brand-narrative backstory. Each topic's event description is provided in the prompt; what the model must additionally bring is the prior knowledge of how this kind of event has historically been framed by RedNote consumers---which colors get hyped, which corporate moves trigger hailstorm-apple-style farming-aid narratives, which holidays anchor seasonal photography posts. Capable consumer simulation requires this anticipation; failing it is exactly the capability gap \benchmark{} measures. Models with broader affective range capture more of the visible part of the discourse but still miss the latent-anchor class.

Table~\ref{tab:universal_misses} samples difficult tail criteria across diverse topics, with a brief gloss of why each is operationally important and what kind of prior knowledge a model would need to bring (beyond the event description) to anticipate it.

\begin{table}[!htbp]
\centering
\caption{Difficult tail criteria sampled from the final benchmark. Rows include zero-coverage criteria and the hailstorm-apple near-universal miss covered only by MiMo-V2.5-Pro. Each is a high-salience reaction observed in the source materials, requiring the model to bring platform-specific prior knowledge (beyond the event brief) to anticipate.}
\label{tab:universal_misses}
\small
\renewcommand{\arraystretch}{1.12}
\begin{tabular}{@{}p{0.05\columnwidth}p{0.20\columnwidth}p{0.22\columnwidth}p{0.45\columnwidth}@{}}
\toprule
\rowcolor{CSBPanel}
\textbf{Sec} & \textbf{Topic} & \textbf{Tail criterion} & \textbf{Why obvious in retrospect} \\
\midrule
\rowcolor{CSBPanel}
\multicolumn{4}{@{}l}{\textbf{\textcolor{CSBInk}{T1. Sentiment flashpoints}}} \\
T1 & First fall-leaf photo of the season & ``Solar-term `Start of Autumn' as anchor'' & Chinese users frame seasonal photography by the 24-solar-term calendar, not by the calendar month. The event description does not surface this anchor. \\
T1 & Outdoor enthusiasts have their own real estate & ``Their `real estate' is the gear wall'' & The viral framing is a specific meme: outdoor gear arranged like home decor. Models pivot to generic outdoor enthusiasm, missing the meme structure. \\
T1 & Aaaaapple files police report & ``Hailstorm-damaged apple farming-help backstory'' & The actual hook is that the brand sells crop-damaged fruit to support farmers. Twelve models discuss the police report but never reach the upstream narrative; interestingly, only MiMo covers this anchor. \\
T1 & Tang Xiang-Yu stand-up tour & ``Single-table / self-seating independence'' & The viral payoff is a specific feminist metaphor: stop waiting to be invited to the table and open one yourself. All 13 models discuss empowerment generically but miss the exact phrase that carries the crowd reaction. \\
\addlinespace[0.2em]
\rowcolor{CSBPanel}
\multicolumn{4}{@{}l}{\textbf{\textcolor{CSBInk}{T2. Emotion keywords}}} \\
T2 & Aaaaapple files police report & ``Altruistic pride (`I bought it to help')'' & First-person consumer pride is the dominant supportive register on this topic. Models default to third-party commentary. \\
T2 & Tang Xiang-Yu stand-up tour & ``Empowerment / catharsis as women'' & The female-empowerment register is the show's whole brand; models cover ``funny'' and ``relatable'' but miss the demographic-anchored emotional payoff. \\
\addlinespace[0.2em]
\rowcolor{CSBPanel}
\multicolumn{4}{@{}l}{\textbf{\textcolor{CSBInk}{T3. Positive aspects}}} \\
T3 & iPhone 17 hands-on & ``Color-option attractiveness'' & Color SKUs are a real RedNote driver of pre-order excitement. Models discuss specs and incrementalism, never the colorways. \\
T3 & Aaaaapple files police report & ``Brand thoughtfulness (packaging / gifting)'' & Defense-of-brand posts emphasize packaging detail and small gifts as evidence of sincerity. Models do not generate this defense vector. \\
\addlinespace[0.2em]
\rowcolor{CSBPanel}
\multicolumn{4}{@{}l}{\textbf{\textcolor{CSBInk}{T4. Negative aspects}}} \\
T4 & iPhone 17 hands-on & ``China-region feature limitations (eSIM / AI / health)'' & A standing concern across every Apple cycle in China; would be obvious to any user but absent from the event description. \\
\bottomrule
\end{tabular}
\end{table}

\paragraph{What characterizes a zero-coverage criterion.}
Three properties recur.
\textbf{(i)}~Anchored in a specific cultural register (solar terms, demographic-empowerment vocabulary, region-policy concerns) that a one-paragraph event description doesn't surface.
\textbf{(ii)}~Often a viral meme template applied to a niche (``X has its own real estate''). Models recognize the template only when the topic explicitly states it.
\textbf{(iii)}~First-person consumer voice (``I bought to help,'' ``I'm proud to support'') --- models default to third-party commentary, not first-person purchase narratives.

\paragraph{Anatomy of a near-universal miss.}
For one 1/13 example, we expose the full 5-field structure. The criterion is well-defined, has positive and negative anchors, and a clear judgment rule---yet only MiMo surfaces the broader brand-story anchor.

\casebox{
\textbf{T1 ``Hailstorm-damaged apple farming-help backstory'' (near-universal miss, 1/13; only MiMo covers)}\\
\emph{Topic.} Aaaaapple Apple Juice files police report.\\
\emph{Definition.} The brand's actual business model is buying hailstorm-damaged (otherwise unsellable) apples from farmers and processing them into juice / chips---this farming-aid backstory is the viral hook for the celebrity endorsement.\\
\emph{Positive examples.} ``Hail-damaged crop apples that would've gone to waste---this brand turns them into juice.''\quad ``It's not just `endorsement,' it's that Lusi found the actual farming-aid niche.''\quad ``Cosmetic apples nobody buys at the wholesale market---now they're useful.''\\
\emph{Negative examples.} ``I support charity products in general'' (generic, no hailstorm-anchor).\quad ``Helping farmers is good'' (generic charity praise).\\
\emph{Judgment rule.} Must mention hailstorm-damaged, ugly-but-edible, or otherwise unsellable produce as the specific farming-aid mechanism.\\[0.4em]
\textbf{What almost every model produced instead.} Twelve generators output generic ``helping farmers is admirable'' / ``charity is good'' phrasings. MiMo is the sole model that retrieves the upstream brand story by writing, ``the farming-aid activity was originally a good thing.'' It still misses the exact hailstorm mechanism, but unlike the other models it pivots from the police-report surface event to the consumer-facing reason people defended the brand. This is what makes the criterion ``obvious in retrospect'': once you know the brand sells hail-damaged apples, half the public discourse makes sense---but the event description in the prompt does not name it.\\[0.3em]
\textbf{Why it matters.} For a marketing application, this is exactly the type of brand-narrative anchor a forecasting system would need to surface. A model that misses this misses the actual reason the topic gained engagement.
}

\paragraph{Operational implication.}
These difficult-tail criteria are exactly the reactions a marketing practitioner needs predicted, from the cultural anchor that drives engagement to the concrete business-model fact that explains why a topic resonated. They define the actual frontier of crowd-reaction reconstruction.

\FloatBarrier

\end{document}